\documentclass[runningheads]{llncs}
\usepackage[T1]{fontenc}
\usepackage{graphicx}
\usepackage{booktabs}
\IfFileExists{ifsym.sty}{\usepackage[misc]{ifsym}}{}

\usepackage{amsmath}
\usepackage{amssymb}
\usepackage{multirow}
\usepackage{xcolor}
\usepackage{listings}
\usepackage{realboxes} 
\usepackage{url}

\definecolor{artelt}{RGB}{31, 119, 180}
\definecolor{dice}{RGB}{174, 199, 232}
\definecolor{cchvae}{RGB}{255, 127, 14}
\definecolor{ppcef}{RGB}{255, 187, 120}
\definecolor{cegp}{RGB}{44, 160, 44}
\definecolor{cadex}{RGB}{152, 223, 138}
\definecolor{sace}{RGB}{214, 39, 40}
\definecolor{ares}{RGB}{196, 156, 148}
\definecolor{globece}{RGB}{140, 86, 75}
\definecolor{globalglance}{RGB}{227, 119, 194}
\definecolor{glance}{RGB}{247, 182, 210}
\definecolor{tcrex}{RGB}{127, 127, 127}
\definecolor{cearm}{RGB}{31, 119, 180}
\definecolor{wach}{RGB}{255, 127, 14}

\definecolor{lightblue}{RGB}{230, 240, 255}
\definecolor{codegreen}{rgb}{0,0.6,0}
\definecolor{codegray}{rgb}{0.5,0.5,0.5}
\definecolor{codepurple}{rgb}{0.58,0,0.82}
\definecolor{backcolour}{rgb}{0.95,0.95,0.92}

\lstdefinestyle{mystyle}{
    backgroundcolor=\color{lightblue},
    commentstyle=\color{codegreen},
    keywordstyle=\color{blue},
    numberstyle=\tiny\color{codegray},
    stringstyle=\color{red},
    basicstyle=\ttfamily\footnotesize,
    breakatwhitespace=false,
    breaklines=true,
    captionpos=b,
    keepspaces=true,
    numbers=left,
    numbersep=4pt,
    showspaces=false,
    showstringspaces=false,
    showtabs=false,
    tabsize=2,
    frame=ltb,
    framerule=0pt,
    keywordstyle={\bfseries \color{blue}}
}
\newcommand{\code}[1]{\Colorbox{lightblue}{\lstinline{#1}}}

\begin{document}

\title{CEL: Comprehensive Counterfactual Explanations Library and Benchmark}
\titlerunning{CEL: Counterfactual Explanations Library and Benchmark}

\author{Oleksii Furman\thanks{O. Furman and Ł. Lenkiewicz contributed equally to this research.} \and 
Łukasz Lenkiewicz$^*$ \and 
Marcel Musiałek \and 
Maciej Zięba}
\authorrunning{O. Furman et al.}
\institute{Wrocław University of Science and Technology, Wrocław, Poland \\
\email{\{oleksii.furman, lukasz.lenkiewicz, maciej.zieba\}@pwr.edu.pl}, \email{279704@student.pwr.edu.pl}}

\maketitle

\begin{abstract}
Counterfactual explanations are a prominent approach in explainable artificial intelligence (xAI), providing actionable guidance on what input changes would alter a model’s prediction to a desired outcome. While early methods primarily focused on minimal feature changes, recent work incorporates additional properties such as sparsity, actionability and plausibility. Despite this progress, fair and systematic evaluation remains challenging. Existing studies often rely on different data splits, predictive models, and evaluation metrics, which limits objective comparison across methods. To fill this gap, we introduce CEL (\textbf{C}ounterfactual \textbf{E}xplanations \textbf{L}ibrary), a unified library and benchmark for counterfactual explanations designed to support consistent implementation and evaluation. CEL includes 18 datasets of varying size and complexity and provides implementations or reimplementations of 14 widely used counterfactual methods. Using this standardized setup, we conduct a comprehensive quantitative comparison across a variety of methods on datasets that differ in size, number, and types of attributes. The evaluation protocol incorporates multiple complementary metrics capturing validity, coverage, sparsity, proximity, and distributional plausibility, including density- and outlier-based measures to assess the realism of generated counterfactuals. To the best of our knowledge, this is the first comprehensive benchmark that systematically evaluates recent counterfactual explanation methods within a unified and reproducible framework. Whereas prior libraries are often outdated, limited in scope, or lack consistent protocols, CEL aims to improve reproducibility, enable fair comparison, and provide a workbench for developing future counterfactual explanation methods.

\keywords{Counterfactual Explanations \and Explainable Artificial Intelligence \and Benchmarking \and Machine Learning}
\end{abstract}

\begin{figure}
    \centering
    \includegraphics[width=\textwidth]{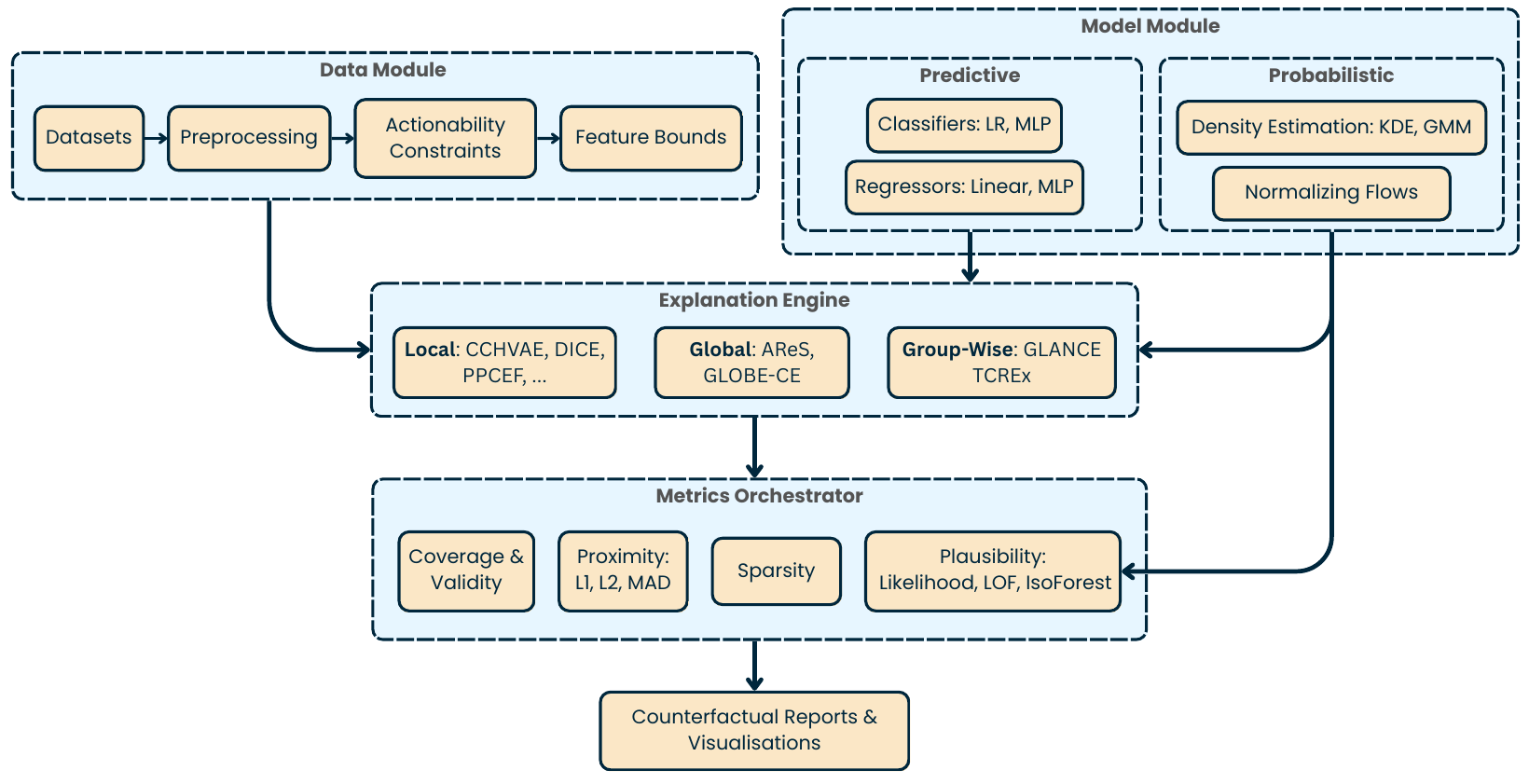}
    \caption{Overview of the CEL architecture. It consists of four interconnected modules: (i) a Data Module handling dataset loading, preprocessing, and constraint specification; (ii) a Model Module providing predictive and probabilistic backbones; (iii) an Explanation Engine supporting local, global, and group-wise counterfactual methods; and (iv) a Metrics Orchestrator computing validity, proximity, sparsity, and plausibility measures. Arrows indicate data flow between components.}
    \label{fig:teaser}
\end{figure}

\section{Introduction}
Machine learning models are increasingly deployed in high-stakes applications, where transparency and accountability are critical. Counterfactual explanations have emerged as a prominent approach in explainable AI, offering actionable insights by identifying minimal changes required to alter a model’s prediction. Early formulations, most notably by Wachter et al.~\cite{wachter2017counterfactual}, focused primarily on two criteria: validity (the counterfactual must achieve the desired prediction) and proximity (the modification from the original instance should be minimal). While intuitive, these constraints alone often lead to unrealistic or impractical explanations.


Subsequent work has expanded the set of desirable properties for counterfactual explanations. In addition to validity and proximity, researchers have emphasized plausibility, requiring counterfactuals to lie in high-density regions of the data distribution; sparsity, encouraging modifications to as few features as possible; and actionability, ensuring that only features that can be feasibly changed in practice are modified \cite{guidotti2024counterfactual}. As a result, modern counterfactual generation methods typically optimize for multiple, and sometimes competing, criteria. This growing set of constraints has significantly increased the complexity of both method design and evaluation. 



Since their introduction, the evaluation of counterfactual methods has remained fundamentally under-specified. For a given instance, multiple valid counterfactuals may exist, each satisfying different subsets of desirable properties \cite{mothilal2020explaining}. Modern methods therefore optimize trade-offs among criteria such as validity, proximity, plausibility, sparsity, robustness, and actionability, requiring a multi-dimensional assessment rather than a single scalar metric \cite{guidotti2024counterfactual}.

In practice, studies use inconsistent evaluation setups: metrics are defined differently, different subsets of properties are emphasized, and experimental protocols vary in data splits, preprocessing, feature encodings, constraints, and the predictive models being explained \cite{de2021framework,pineau2021improving}. Because counterfactuals are inherently model-dependent, such variation can substantially affect feasibility and quality, so observed differences may reflect experimental configuration rather than methodological advances. This lack of standardization makes fair comparison difficult, yields unstable method rankings, and hinders reliable measurement of progress.

To address this fragmentation, we introduce CEL, a unified benchmark and reproducible evaluation framework. Rather than serving solely as an implementation library, CEL establishes a controlled protocol that standardizes datasets, predictive backbones, preprocessing pipelines, feature constraints, and evaluation metrics, enabling systematic and fair comparison across counterfactual paradigms. Figure~\ref{fig:teaser} overviews the CEL architecture, connecting data management, predictive and generative modeling, explanation generation, and standardized evaluation.

CEL includes 18 curated datasets spanning diverse domains and 14 widely used counterfactual generation methods implemented or reimplemented within a common interface. All methods are evaluated under harmonized training splits, shared predictive models, and a consistent metric taxonomy. The library is publicly available via \code{pip install ce-library} and the source code is hosted on https://github.com/ofurman/counterfactuals

\smallskip
\noindent
Our main contributions are:

\begin{itemize}
\item \textbf{A controlled evaluation protocol} for counterfactual explanations that standardizes datasets, preprocessing, predictive backbones, constraint handling, and metric definitions, enabling fair and reproducible comparison.
\item \textbf{A benchmark including} 18 datasets and 14 counterfactual generation methods implemented within a unified framework, supporting evaluation across local, global, and group-wise paradigms.
\item \textbf{An open-source programming library} designed to facilitate future method integration, transparent reporting, and community-driven benchmarking.
\end{itemize}

\section{Related Works}

Research on counterfactual explanations has led to several libraries and benchmarking efforts designed to support implementation and evaluation. However, despite these contributions, a fully controlled and protocol-standardized benchmark for counterfactual explanations is still lacking.

CARLA \cite{pawelczyk2021carla} is among the earliest and most widely adopted benchmarking libraries for counterfactual explanations and algorithmic recourse, providing multiple generation methods, integrated evaluation measures, and several datasets within a unified Python framework. While it greatly improved accessibility and comparability, it does not enforce a fully fixed experimental protocol across predictive backbones, preprocessing choices, and constraint configurations.

More recently, RobustX \cite{jiang2025robustx} targets the robustness of counterfactual explanations under model and data perturbations, offering standardized tools for this property. While valuable for analyzing stability, its scope is narrow and does not cover multiple counterfactual paradigms under a unified experimental protocol.

In contrast to prior libraries, CEL emphasizes protocol-level control across datasets, predictive backbones, preprocessing steps, constraint handling, and metric definitions, enabling systematic and fair comparison across local, global, and group-wise counterfactual paradigms.

\section{CEL: Counterfactual Explanations Library}

CEL is built from modular components that separate models, datasets, counterfactual methods, and evaluation metrics. This structure allows flexible composition and simplifies extension with new methods, datasets, and metrics, while typed interfaces ensure consistent data handling.

\subsection{Architecture and Design Principles}

CEL enforces a strict separation of concerns, so that data preprocessing, model training, explanation generation, and evaluation operate as decoupled yet interoperable components. It employs a \textbf{configuration-driven} workflow, letting researchers define experiments---model hyperparameters, dataset constraints, and metric selection---via hierarchical configuration files rather than hard-coded scripts.

\subsection{Unified Method Interfaces}

A key challenge in benchmarking is the heterogeneity of counterfactual algorithms. CEL unifies them under a common abstraction, \code{BaseCounterfactualMethod}, which defines a standardized interface for explanation generation. All methods return a structured \code{ExplanationResult}, enabling consistent evaluation and direct comparison across approaches.

To support diverse explanation paradigms, CEL adopts a Mixin-based design that unifies local, global, and group-wise methods under a common interface, enabling consistent integration and evaluation.


\subsection{Data Management and Preprocessing}

Valid counterfactual evaluation requires rigorous data handling to ensuring that generated instances remain within valid feature domains. CEL implements a robust \code{DatasetBase} architecture that manages feature metadata, including actionability constraints, mutability, and feature bounds.

The library features a flexible \code{PreprocessingPipeline}, allowing users to chain transformations such as scalers (Standard, MinMax, Robust) and encoders (OneHot, Ordinal). Crucially, the pipeline supports precise inverse transformations, ensuring that counterfactuals generated in processed space can be accurately mapped back to the original input space for evaluation.

For methods requiring continuous density estimation (e.g., normalizing flows), CEL includes a specialized \code{Dequantization} module. This module converts discrete categorical features into continuous values using variational or noise-based dequantizers, bridging the gap between discrete tabular data and continuous generative models.

\subsection{Predictive and Generative Backbones}

CEL supports a wide range of model architectures via a unified \code{PytorchBase} interface, categorized into task-specific mixins.

\textbf{Discriminative Models.} To ensure broad compatibility, the library implements \code{ClassifierPytorchMixin} and \code{RegressionPytorchMixin}. Implemented architectures range from standard Multi-Layer Perceptrons (MLP) and Logistic Regression to advanced Neural Oblivious Decision Ensembles (NODE), allowing researchers to test counterfactual generation against models of varying complexity and interpretability and easily extend their code with new implementations.

\textbf{Generative Models.} Some of modern counterfactual methods and metrics rely on density estimation to ensure the plausibility of generated explanations. CEL integrates a \code{GenerativePytorchMixin} that standardizes likelihood estimation and sampling interfaces. The library provides implementations of state-of-the-art Normalizing Flows, including Masked Autoregressive Flow (MAF) \cite{papamakarios2017masked}, along with RealNVP and NICE. Additionally, it supports Kernel Density Estimation (KDE) and Gaussian Mixture Models (GMM) for baseline comparisons. These models allow users to estimate the likelihood of generated counterfactuals, a key component in assessing plausibility.

\subsection{Evaluation Orchestration}

CEL introduces a \code{MetricsOrchestrator} which allows to use existing metrics or define custom ones. This component leverages a registry-based system to dynamically instantiate and compute metrics defined in the experiment configuration. The orchestrator handles input validation and compute a comprehensive suite of metrics covering validity, proximity, sparsity, and density-based plausibility scores. Metrics are detailed in the supplementary material.

\vspace{1em}

CEL modular workflow is demonstrated in Listing \ref{lst:code_snippet}, which illustrates how the Dataset, Model, Method, and MetricsOrchestrator components interact. As shown, the framework abstracts the complexity of training and optimization, allowing users to generate and evaluate counterfactuals with minimal boilerplate code.

\begin{lstlisting}[float=tp, belowskip=-1em, language=Python, backgroundcolor=\color{lightblue}, caption={Code snippet showing CEL modular workflow.}, label={lst:code_snippet}]
from cel.datasets import FileDataset
from cel.models.classifiers import MLPClassifier
from cel.cf_methods.local_methods import PPCEF
from cel.metrics import MetricsOrchestrator

# 1. Configuration-driven Data Loading
dataset = FileDataset(config_path="dataset.yaml")

# 2. Model Training (Unified PytorchBase Interface)
classifier = MLPClassifier(...)
classifier.fit(dataset.train_dataloader())

# 3. Counterfactual Generation (LocalMixin)
method = PPCEF(classifier, gen_model=flow_model, ...)
result = method.explain_dataloader(dataset.test_dataloader())

# 4. Standardized Evaluation
metrics = MetricsOrchestrator(conf_path="default.yaml")
scores = metrics.compute(result)
\end{lstlisting}

\section{Benchmark}

A good counterfactual benchmark rests on three principles \cite{pawelczyk2021carla,de2021framework}. First, a \textbf{standardized evaluation protocol}---a fixed suite of diverse datasets, pre-defined predictive models, and consistent preprocessing pipelines---eliminates experimental variation as a confounding factor, a common issue in prior studies \cite{pineau2021improving}. Second, \textbf{comprehensiveness}: broad coverage of methods spanning local, group-wise, and global paradigms, evaluated with multi-faceted metrics that capture validity, proximity, sparsity, and plausibility \cite{guidotti2024counterfactual}. Third, \textbf{reproducibility and extensibility}: an open-source design that lets researchers replicate results and integrate their own methods, datasets, and metrics. CEL is built around these principles.

\subsection{Datasets}
CEL includes 18 pre-configured datasets covering classification (13) and regression (5) tasks, spanning domains such as finance and social sciences to test methods across different distributions and complexities. Table~\ref{tab:datasets_complete} summarizes their characteristics.

\begin{table}[!ht]
\centering
\footnotesize 
\setlength{\tabcolsep}{2pt} 
\caption{Datasets included in CEL with feature types, distribution and references.}
\label{tab:datasets_complete}
\begin{tabular}{l l r r r r r}
\toprule
\textbf{Dataset} & \textbf{Task} & \textbf{$N$} & \textbf{Cat} & \textbf{Num} & \textbf{$C$} & \textbf{Label Distribution} \\
\midrule
Adult Census \cite{adult} & Clf. & 32,561 & 8 & 4 & 2 & 0: 75.9\%, 1: 24.1\% \\
Audit \cite{audit} & Clf. & 775 & 0 & 23 & 2 & 0: 60.6\%, 1: 39.4\% \\
Bank Marketing \cite{moro2014bank} & Clf. & 40,004 & 9 & 7 & 2 & 0: 88.3\%, 1: 11.7\% \\
Blobs & Clf. & 1,500 & 0 & 2 & 3 & 0, 1, 2: 33.3\% each \\
Credit Default \cite{credit_default_data} & Clf. & 30,000 & 9 & 14 & 2 & 0: 77.9\%, 1: 22.1\% \\
Digits \cite{digits} & Clf. & 1,797 & 0 & 64 & 10 & Balanced ($\sim$10\% each) \\
German Credit \cite{german_credit_data} & Clf. & 1,000 & 11 & 7 & 2 & 0: 70.0\%, 1: 30.0\% \\
GMC \cite{pawelczyk2020learning} & Clf. & 16,714 & 3 & 7 & 2 & 0: 50.0\%, 1: 50.0\% \\
HELOC \cite{heloc} & Clf. & 10,459 & 0 & 23 & 2 & 0: 52.2\%, 1: 47.8\% \\
Law \cite{law} & Clf. & 2,216 & 2 & 3 & 2 & 0: 50.0\%, 1: 50.0\% \\
Lending Club \cite{jagtiani2019roles} & Clf. & 93,888 & 4 & 8 & 2 & 0: 29.6\%, 1: 70.4\% \\
Moons & Clf. & 1,024 & 0 & 2 & 2 & 0: 50.0\%, 1: 50.0\% \\
Wine \cite{wine} & Clf. & 178 & 0 & 13 & 3 & 0: 33.1, 1: 39.9, 2: 27\% \\
\midrule
Synthetic & Regr. & 1,000 & 0 & 2 & Cont. & [0, 1] \\
Concrete \cite{concrete_compressive_strength_data} & Regr. & 1,030 & 0 & 8 & Cont. & [0, 1]\\
Diabetes \cite{diabetes_data} & Regr. & 442 & 0 & 10 & Cont. & [0, 1] \\
Yacht \cite{yacht_hydrodynamics_data} & Regr. & 308 & 0 & 6 & Cont. & [0, 1] \\
SCM20D \cite{scm20d_data} & Regr. & 8,966 & 0 & 61 & Cont. (16) & $[0, 1]^{16}$ \\
\bottomrule
\end{tabular}
\end{table}


Consistent preprocessing is applied across all datasets. Continuous features are scaled to the $[0, 1]$ range using Min-Max normalization. Categorical features are handled based on the method's requirements For discriminative models and CF methods, categorical features are maintained in their original encoded format. to facilitate training of density estimator we apply variational dequantization to discrete features, mapping them into a continuous latent space  while maintaining the discrete structure during inverse transformations.

\subsection{Models}
CEL integrates distinct types of predictive and generative models implemented in PyTorch, which are essential for generating and evaluating counterfactual explanations. To ensure rigorous evaluation, CEL implements several predictors that we used in our benchmark. \textbf{Classification Models:} The library includes implementations of Multi-Layer Perceptrons (MLP), Logistic Regression. \textbf{Regression Models:} For continuous target variables, CEL provides Linear Regression and MLP Regressor models. \textbf{Density Estimator:} To support plausibility evaluation and methods that rely on density estimation we utilise Masked Autoregressive Flow, accordingly to Wielopolski et al.~\cite{wielopolski2024probabilistically}.

\subsection{Methods}
The library implements 14 counterfactual explanation methods, categorized into local, global, and group-wise approaches. All methods adhere to a common interface, facilitating direct comparison.

\begin{table}[h!]
\centering
\caption{Counterfactual explanation methods included in the CEL benchmark.}
\label{tab:methods}
\begin{tabular}{l l l}
\toprule
\textbf{Category} & \textbf{Method} & \textbf{Citation} \\
\midrule
\multirow{9}{*}{Local} & Wachter & \cite{wachter2017counterfactual} \\
& Artelt & \cite{artelt2020convex} \\
& DiCE & \cite{mothilal2020explaining} \\
& CCHVAE & \cite{karimi2020model} \\
& PPCEF & \cite{wielopolski2024probabilistically} \\
& CEM & \cite{DhurandharCLTTS18} \\
& CEGP & \cite{LooverenK21} \\
& CADEX & \cite{moore2019explaining} \\
& SACE & \cite{keane2020good} \\
& CEARM & \cite{spooner2021counterfactual} \\
\midrule
\multirow{2}{*}{Global} & AReS & \cite{rawal2020beyond} \\
& GLOBE-CE & \cite{ley2023globe} \\
\midrule
\multirow{2}{*}{Group-wise} & GLANCE & \cite{kavouras2024glance} \\
& T-CREx & \cite{bewley2024tcrex} \\
\bottomrule
\end{tabular}
\end{table}

\subsubsection{Local Methods}
Local methods generate explanations for individual instances. CEL includes classic optimization-based approaches such as Wachter et al. \cite{wachter2017counterfactual} and DiCE \cite{mothilal2020explaining}, which solve optimization problems to generate counterfactuals. The library also supports several methods that focus on plausibility via density-based constraints and generative models, including the method of Artelt et al.~\cite{artelt2020convex} (Artelt); CCHVAE \cite{karimi2020model}, which leverages variational autoencoders for manifold-constrained generation; PPCEF \cite{wielopolski2024probabilistically}, a flow-based approach with an explicit probabilistic formulation; and CEGP \cite{LooverenK21}, which guides perturbations toward prototype-based counterfactuals. CEM \cite{DhurandharCLTTS18} uses autoencoders to verify the plausibility of perturbed instances. The case-based method SACE \cite{keane2020good} and CADEX \cite{moore2019explaining} (constrained adversarial examples) are also included. Additionally, CEL implements CEARM \cite{spooner2021counterfactual}, a Bayesian optimization-based method designed for regression models with a globally convergent search algorithm.

\subsubsection{Global Methods}
Global methods provide dataset-level insights. CEL implements AReS \cite{rawal2020beyond}, which generates global recourse rules, and GLOBE-CE \cite{ley2023globe}, which learns global translation directions. Additionally, we include GLANCE \cite{kavouras2024glance} in our comparison by configuring it with a single group, effectively adapting this group-wise method for a global evaluation. These methods are essential for understanding high-level model behavior and identifying systemic biases.

\subsubsection{Group-wise Methods}
Bridging the gap between local and global, group-wise methods generate explanations for subgroups of data. CEL includes GLANCE \cite{kavouras2024glance} and TCREX \cite{bewley2024tcrex}, which allow users to inspect counterfactuals for specific clusters or demographic groups.

\subsection{Metrics}
\label{sec:metrics}
To ensure a holistic evaluation, CEL provides a comprehensive suite of metrics covering key counterfactual properties: Validity, Proximity, Sparsity, Plausibility. To ensure rigor evaluation we selected such metrics to evaluate each property:
\begin{itemize}
    \item \textbf{Validity:} Measures the success rate of generated counterfactuals in achieving the target prediction. For continuous targets, validity is measured as the mean absolute error (MAE) between the predicted value of the counterfactual and the target value. 
    \item \textbf{Proximity:} Quantifies the distance between the original instance and the counterfactual using \textit{Euclidean} distance for continuous features. For mixed data we apply combined \textit{Euclidean-Hamming} distance, weighted proportionally by number of continuous and discrete features.
    \item \textbf{Sparsity:} Evaluates the fraction of modified features, encouraging simpler explanations.
    \item \textbf{Plausibility:} Assesses how well counterfactuals fit the data distribution using Log-Likelihood from density estimation model. We use MAF \cite{papamakarios2017masked} for this benchmark, due to its superiority verified by Wielopolski et al.~\cite{wielopolski2024probabilistically}.
\end{itemize}
This evaluation ensures that methods are not just optimized for a single criterion (e.g., validity) but produce high-quality, actionable, and realistic explanations. We include a detailed explanation of all metrics in the supplementary material.

\begin{figure}[ht]
    \centering
    \includegraphics[width=\textwidth]{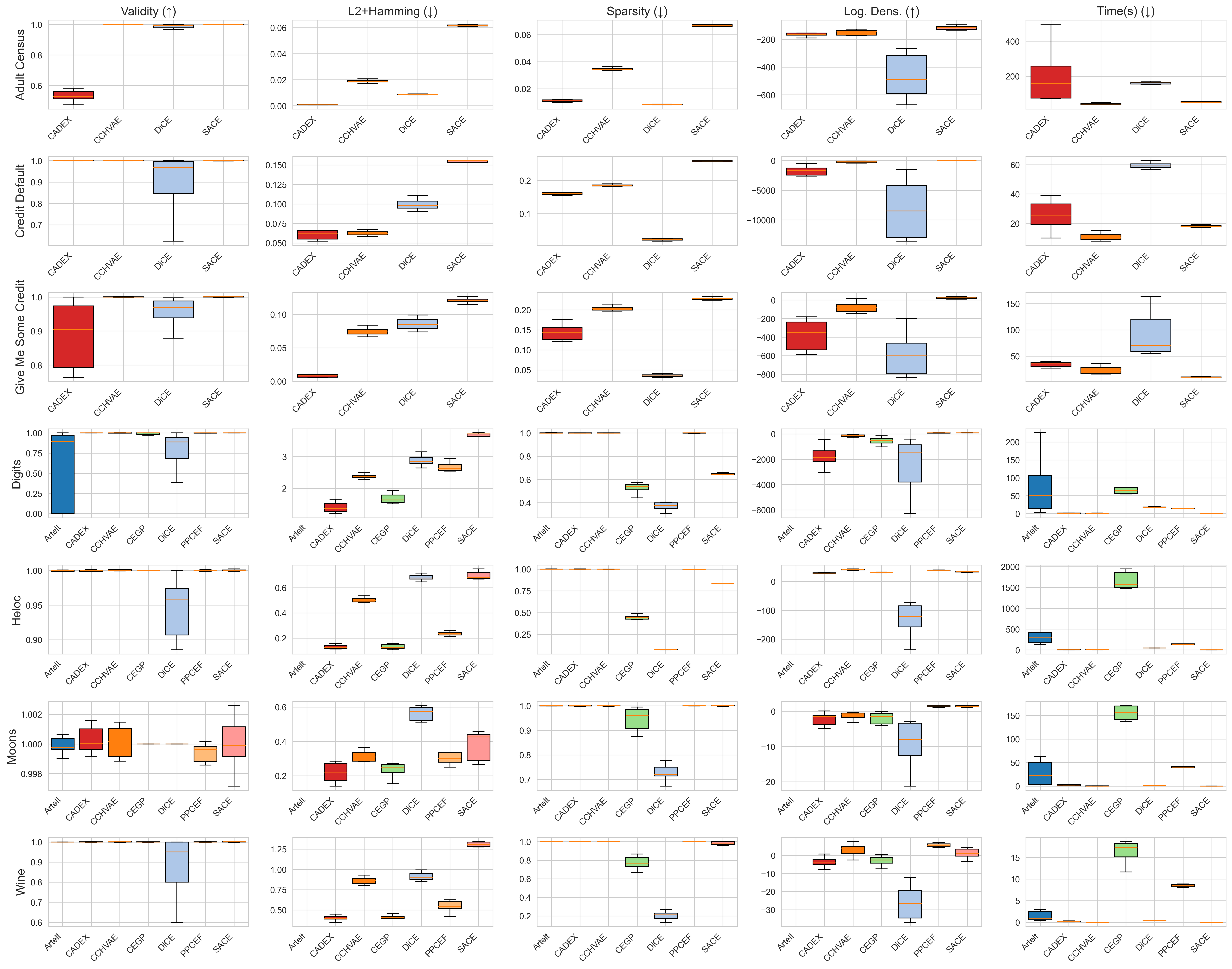}
    \caption{Performance of local counterfactual methods across seven classification datasets. Rows represent datasets and columns represent evaluation metrics (Validity, Euclidean-Hamming Distance, Sparsity, Log-Density, Computation Time). Each boxplot shows the distribution of metric values across methods: \textcolor{artelt}{\textbf{Artelt}} (blue), \textcolor{dice}{\textbf{DiCE}} (light blue), \textcolor{cchvae}{\textbf{CCHVAE}} (orange), \textcolor{ppcef}{\textbf{PPCEF}} (light orange), \textcolor{cegp}{\textbf{CEGP}} (green), \textcolor{cadex}{\textbf{CADEX}} (red), and \textcolor{glance}{\textbf{SACE}} (pink).}
    \label{fig:local_methods}
\end{figure}

\subsection{Experimental Setup}
Our evaluation systematically compares counterfactual methods across local, global, and group-wise paradigms under identical conditions, isolating methodological differences from confounding factors. We evaluate all methods on 18 datasets of varying size, dimensionality, and feature composition, using two predictive backbones per task type, and assess validity, proximity, sparsity, and plausibility to reveal both absolute performance and the trade-offs and failure modes that emerge across diverse settings.

We adopt a standardized protocol using 5-fold cross-validation. For each fold we train the discriminative model on the training split, then fit a generative density estimator on data labeled by that model's predictions, so conditional density is estimated against the appropriate class distribution. We model class-conditional densities with masked autoregressive flows (MAF) \cite{papamakarios2017masked}, which flexibly estimate complex distributions while preserving tractable likelihood computation.

For each dataset we define a target class and explain test instances whose prediction differs from it. For multiclass datasets we restrict evaluation to a binary setting between a selected class pair, aligning generation and density estimation with a well-defined decision boundary and enabling consistent comparison across datasets.

For regression tasks, we set the problem to increasing the predicted value. The desired target value is defined as the original prediction increased by 20\% of the overall range of the target variable. We measure validity as the Mean Absolute Error (MAE) between the counterfactual's prediction and this target value.

\subsection{Results}

In this section, we present representative results from our benchmark. The complete evaluation across all datasets and models is provided in the supplementary material. Our benchmark reports multiple metrics capturing different properties of counterfactual explanations, including validity, proximity, and plausibility. This multi-dimensional evaluation enables analysis of trade-offs between competing objectives. For presentation, we selected a representative subset of datasets that covers the key data characteristics encountered in our full benchmark: purely numerical features (Digits, Wine, HELOC, Moons), mixed categorical and numerical features (Give Me Some Credit, Credit Default, Adult Census), and regression targets (Concrete, Diabetes, Synthetic). All reported results are averaged across both Logistic Regression and MLP classifiers to ensure robustness across different model architectures.

\subsubsection{Local Methods}

Figure~\ref{fig:local_methods} presents the performance of local methods across seven datasets. Most of the methods achieve perfect validity across all datasets, while CADEX shows moderate performance for mixed data and Artelt demonstrates the highest variance. The consistency of CCHVAE and PPCEF makes them suitable for applications where validity is the primary concern. CADEX generates the closest counterfactuals, followed by PPCEF and CCHVAE. SACE produces significantly more distant counterfactuals, suggesting counterfactual explanations with higher cost. DICE achieves the lowest sparsity, followed by CEGP and CADEX. PPCEF and CCHVAE demonstrate the high log-density, indicating its counterfactuals align well with the data manifold. DICE and CADEX show moderate plausibility with lower densities, suggesting their counterfactuals may be less realistic. CCHVAE is the fastest method (11.70 s on average), followed by CADEX, while CEGP is the slowest (474.4 s), making it impractical for large-scale applications.

\begin{figure}[t]
    \centering
    \includegraphics[width=1.0\textwidth]{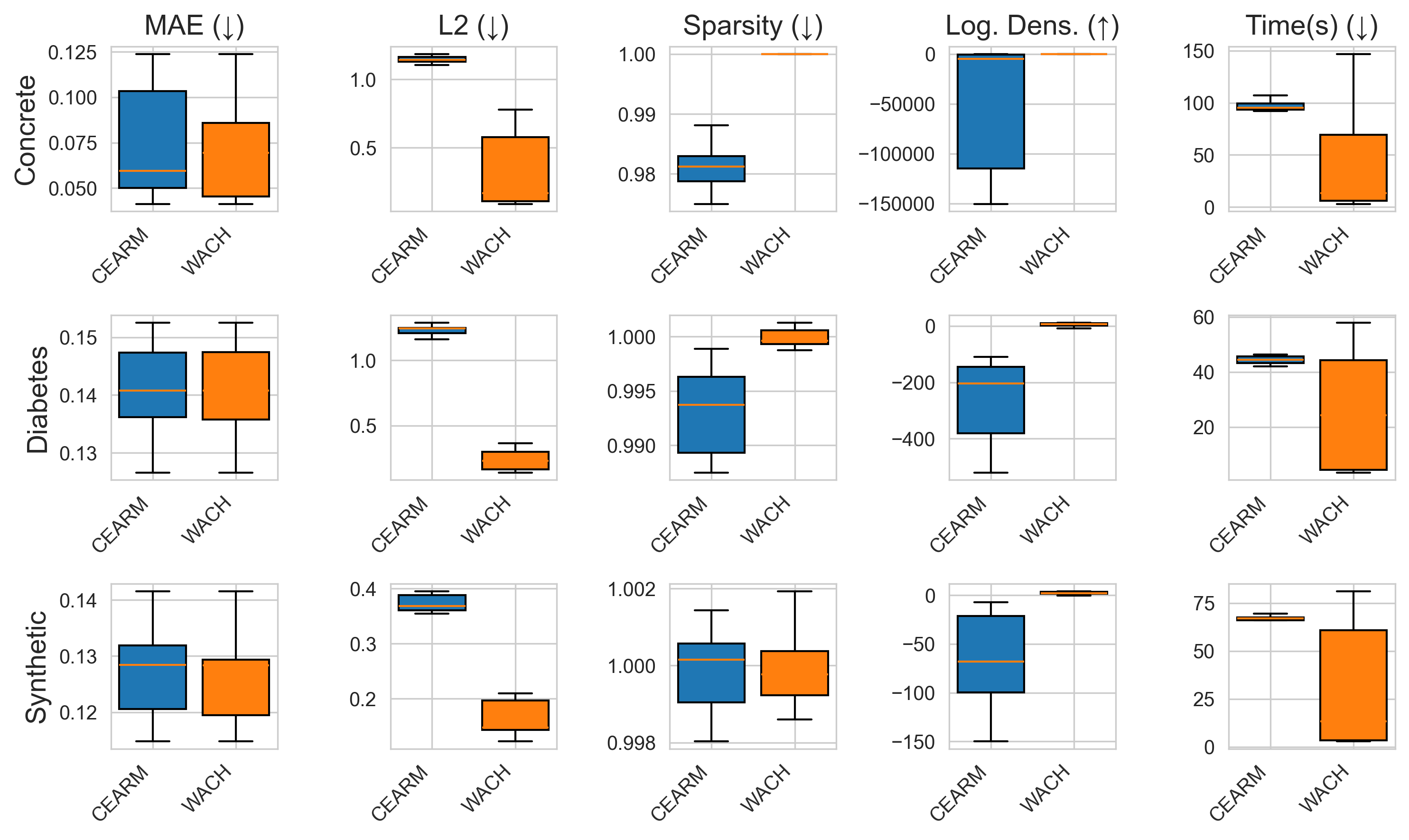}
    \caption{Performance of counterfactual methods on regression tasks. Rows represent regression datasets and columns represent metrics (MAE, Sparsity, Log-Density, Computation Time). Methods: \textcolor{cearm}{\textbf{CEARM}} (blue), \textcolor{wach}{\textbf{WACH}} (orange).}
    \label{fig:regression_methods}
\end{figure}

\paragraph{Regression Methods}
For regression tasks (Figure~\ref{fig:regression_methods}), where validity is measured by Mean Absolute Error (MAE), WACH consistently outperforms CEARM. While both methods achieve comparable accuracy (MAE $\sim$0.1) and high sparsity, WACH produces significantly more plausible counterfactuals (positive log-density of 8.50) and is nearly 9$\times$ faster (7.4s vs. 69.5s). This makes WACH the clear preferred method for regression.

\subsubsection{Global Methods}

\begin{figure}[t]
    \centering
    \includegraphics[width=0.6\textwidth]{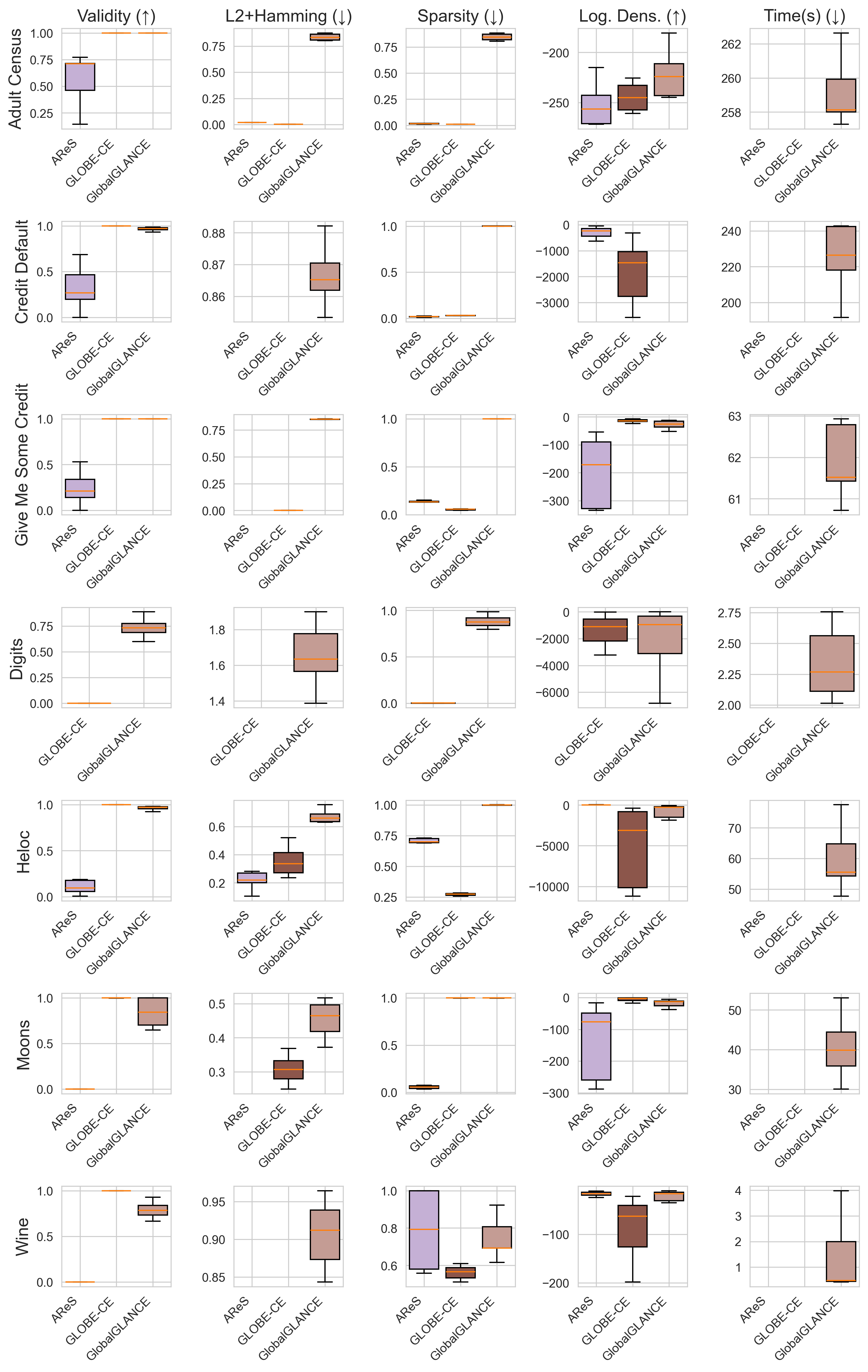}
    \caption{Performance of global counterfactual methods. Rows represent datasets and columns represent metrics. Methods: \textcolor{globalglance}{\textbf{AReS}} (pink), \textcolor{globece}{\textbf{GLOBE-CE}} (dark brown), \textcolor{ares}{\textbf{GlobalGLANCE}} (brown).}
    \label{fig:global_methods}
\end{figure}

Global methods provide dataset-level insights by learning transformations applicable to entire classes. Figure~\ref{fig:global_methods} compares AReS, GLOBE-CE, and GLANCE (with 1 group), aggregated across Logistic Regression and MLP models. GLOBE-CE and GLANCE achieve perfect or near-perfect validity while AReS shows only moderate success rates: GLOBE-CE and GLANCE aggregate explanations by direction with varying magnitude, whereas AReS seeks a single universal fixed change. AReS generates closer counterfactuals when successful but modifies more features, while GLANCE produces more distant counterfactuals with higher sparsity. AReS shows the lowest log-density, though all methods exhibit high variance, reflecting the difficulty of maintaining distributional alignment across diverse instances with a single global transformation.

\subsubsection{Group-wise Methods}

\begin{figure}[t]
    \centering
    \includegraphics[width=0.6\textwidth]{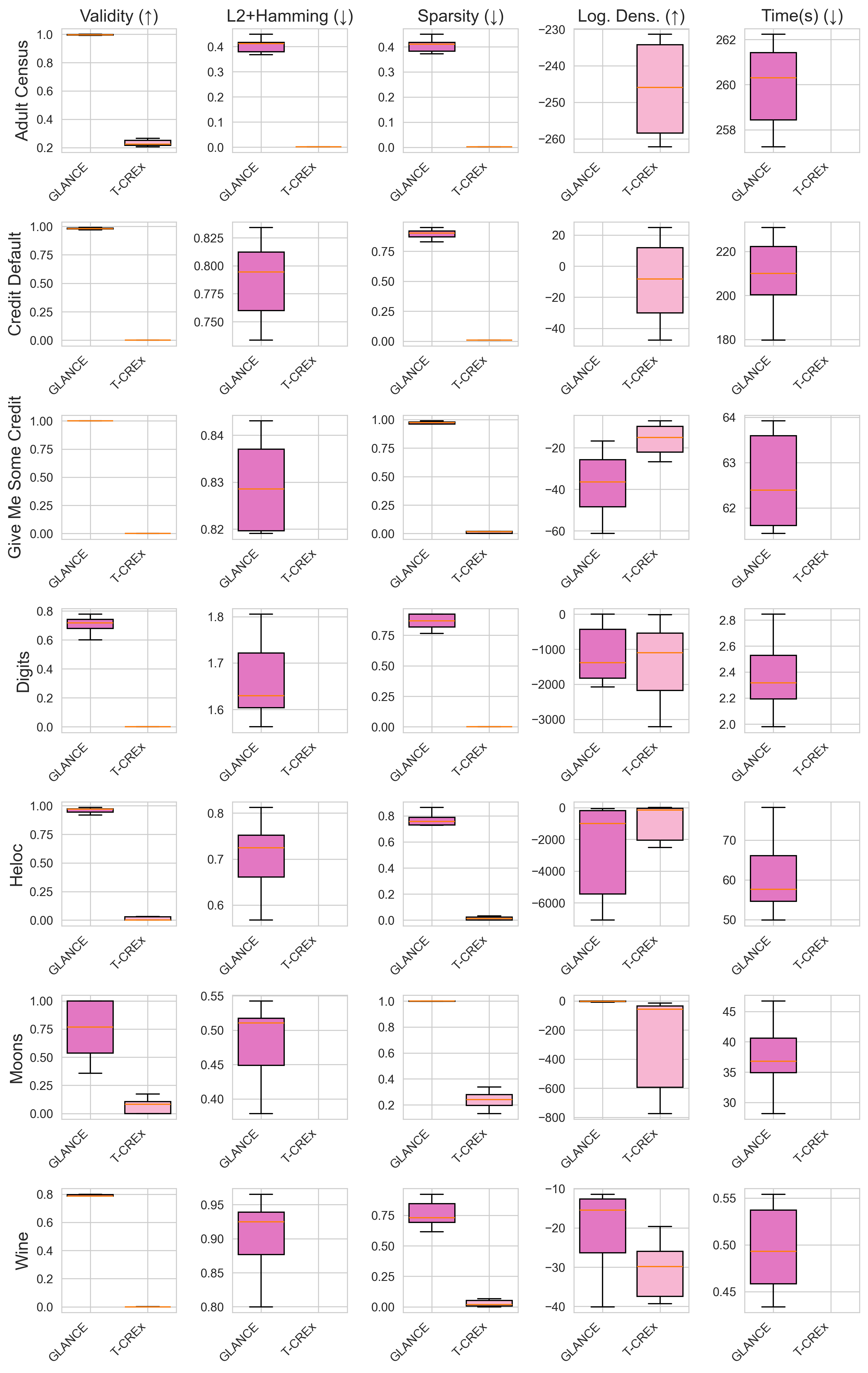}
    \caption{Performance of group-wise counterfactual methods. Rows represent datasets and columns represent metrics. Methods: \textcolor{globalglance}{\textbf{GLANCE}} (dark pink), \textcolor{glance}{\textbf{T-CREx}} (light pink).}
    \label{fig:group_methods}
\end{figure}

Group-wise methods bridge the gap between local and global approaches by generating explanations for subgroups. Figure~\ref{fig:group_methods} presents results for GLANCE and T-CREx, averaged across Logistic Regression and MLP classifiers. This comparison reveals how these methods balance the personalization of local approaches with the efficiency of global methods.

GLANCE consistently achieves higher validity, but T-CREx, when applicable, produces closer and more plausible counterfactuals, however with very low success rates. This highlights a clear trade-off between generating effective and minimally disruptive explanations for subgroups.

Compared to global methods like AReS, group-wise approaches mitigate \textit{validity collapse}. Global methods attempt to find a single recourse rule for the entire population, often resulting in poor generalization (e.g., AReS yields only $0.17$ validity on Adult). By constraining the scope to subgroups, GLANCE recovers the high validity typically associated with local methods while maintaining the interpretability of broad rules. Furthermore, subgroup targeting allows for significantly higher sparsity than global baselines, as the recourse rule need not accommodate the variance of the entire dataset.

Remarkably, GLANCE matches the high validity of local methods despite this harder constraint, though local methods retain superior manifold adherence (lower LOF): group-wise shifts inevitably push peripheral group members into lower-density regions to satisfy the validity constraint for the cluster center. Overall, group-wise counterfactuals offer an effective semi-global explanation, avoiding the noise of local methods and the ineffectiveness of purely global rules.

\section{Conclusions}

We introduced CEL, a unified benchmark and open-source library for standardized evaluation of counterfactual explanation methods. By harmonizing datasets, predictive backbones, preprocessing, and metrics, CEL enables fair and reproducible comparison across local, global, and group-wise paradigms, a significant step beyond the fragmented practices common in the field.

Our benchmark of 14 methods on 18 datasets reveals that no single method uniformly dominates; each excels on some quality dimensions at the expense of others. Among local methods, PPCEF and CCHVAE achieve near-perfect validity with strong plausibility, while CADEX and Wachter produce closer counterfactuals that may fall in low-density regions; DiCE and CEGP yield the sparsest explanations, and SACE delivers the strongest manifold adherence at higher proximity, with computational cost ranging from under 1s (Wachter) to over 470s (CEGP). At the global level, GLOBE-CE achieves minimal, high-validity perturbations whereas AReS frequently fails. Group-wise GLANCE recovers local-level validity with interpretable subgroup rules, while TCREx favors minimal perturbation over effectiveness. For regression, Wachter dominates across all metrics.

These findings underscore that method selection should be guided by application-specific requirements. CEL is designed to be extensible, and we encourage the community to contribute new methods, datasets, and metrics. Promising future directions include evaluating counterfactuals for fairness and robustness, developing user-centric trade-off optimization, and extending to more complex data modalities. We believe CEL will help accelerate progress toward more reliable and trustworthy explainable AI.




\bibliographystyle{splncs04}
\bibliography{sample-base}


\end{document}


\title{Supplementary Material\\CEL: Comprehensive Counterfactual Explanations Library and Benchmark}
\titlerunning{Supplementary Material: CEL}

\author{Author information scrubbed for double-blind reviewing}
\institute{}

\maketitle

\appendix

\section{Evaluation Metrics in CEL}
\label{app:metrics}
To provide a comprehensive assessment of counterfactual methods, CEL implements a broad evaluation suite. We categorize these metrics into three primary dimensions: basic performance, proximity (cost) and plausibility.
\subsection{Basic Metrics}
These metrics assess the fundamental success and efficiency of the counterfactual generation process.
\begin{itemize}
    \item \textbf{Coverage:} The proportion of instances in the test set for which the method successfully returned a counterfactual candidate.
    \item \textbf{Validity:} The proportion of generated counterfactuals that successfully altered the model's prediction to the desired target class $y_{target}$.
    \item \textbf{Regression Validity:} For continuous targets, validity is measured as the mean absolute error (MAE) between the predicted value of the counterfactual and the desired target value.
    \item \textbf{Actionability:} The proportion of instances where the generated counterfactual is identical to the original input (lower is better, as it indicates a failure to provide a meaningful path to recourse).
    \item \textbf{Sparsity:} The average proportion of features modified to achieve the counterfactual. For an input $x$ and counterfactual $x'$, it is defined as:
    \begin{equation}
        \text{Sparsity}(x, x') = \frac{1}{d} \sum_{i=1}^{d} \mathbb{I}(x_i \neq x'_i)
    \end{equation}
\end{itemize}
\subsection{Proximity and Cost}
Proximity metrics measure the distance between the factual instance $x$ and the counterfactual $x'$, representing the "effort" required by the user to achieve recourse. CEL handles mixed-type data by calculating combined distances:
\begin{equation}
    D_{total}(x, x') = \alpha \cdot D_{cont}(x, x') + (1 - \alpha) \cdot D_{cat}(x, x')
\end{equation}
where $\alpha$ is the ratio of continuous features. The library supports:
\begin{itemize}
    \item \textbf{Continuous Distances ($D_{cont}$):} Euclidean ($L_2$), Manhattan ($L_1$), and Median Absolute Deviation (MAD) normalized distance.
    \item \textbf{Categorical Distances ($D_{cat}$):} Hamming distance and Jaccard distance.
\end{itemize}
\subsection{Plausibility and Realism}
Plausibility metrics evaluate whether the counterfactuals lie within the data manifold, ensuring they are not just adversarial perturbations but realistic instances.
\begin{itemize}
    \item \textbf{Log-Density:} The average log-likelihood of counterfactuals estimated by a pre-trained generative model (e.g., Masked Autoregressive Flow). Higher values indicate better alignment with the training distribution.
    \item \textbf{Probabilistic Plausibility:} The fraction of counterfactuals whose log-likelihood exceeds a specified threshold $\tau$ (e.g., the median log-likelihood of the training set) proposed by Wielopolski et al.~\cite{wielopolski2024probabilistically}.
    \item \textbf{LOF/Isolation Forest Scores:} Outlier-based measures where counterfactuals are compared against the training set $X_{train}$ using Local Outlier Factor or Isolation Forest. These identify whether the recourse path leads to "out-of-distribution" regions.
\end{itemize}

\section{Hyperparameter Configurations}
\label{app:hyperparameters}

To provide complete transparency and support reproducibility, this section details the hyperparameters used for all discriminative models, generative models, and counterfactual explanation methods in our benchmark.

\subsection{Discriminative Models}

The predictive models serve as the targets for explanation. We standardize the training procedure across all architectures: epochs: 5,000, learning rate: $0.001$, batch size: 128, and early stopping patience: 300 epochs.

\begin{itemize}
    \item \textbf{MLP:} 2 hidden layers of 256 units each.
    \item \textbf{Logistic Regression:} Standard implementation without additional regularization penalties beyond those inherent in the Adam optimizer.
\end{itemize}

\subsection{Density Estimation Models}

Masked Autoregressive Flow Parameters is used for density-based plausibility metrics and manifold-constrained search. Training hyperparameters include: epochs: 2,000, learning rate: $0.003$, batch size: 1,024. MAF architecture: 8 layers, 4 blocks/layer, 16 hidden features.









\section{Full Results}
\label{app:full_results}

This appendix presents the complete quantitative results of our benchmark evaluation. The main text reports representative examples on selected datasets; here we provide the full set of results across all datasets, methods, and predictive models. All reported values are averaged over 5-fold cross-validation, with standard deviations indicated by $\pm$ notation. Bold values denote the best result for each metric within a given dataset and model combination.

For each experimental configuration, we report a combined metrics table that includes coverage, validity, sparsity, probabilistic plausibility, log-density, LOF score, Isolation Forest score, proximity (L2-Hamming for mixed-type datasets, L2 for numerical datasets), and wall-clock computation time. These tables provide a multi-dimensional view of each method's performance, capturing trade-offs between validity, proximity, plausibility, and computational cost.

Results are organized by counterfactual paradigm and feature type as follows.

\subsection{Local Methods on Mixed-Type Datasets}

This subsection reports results for local counterfactual methods evaluated on seven classification datasets containing both categorical and numerical features: Adult Census, Bank Marketing, Credit Default, German Credit, Give Me Some Credit, Law, and Lending Club. Due to the presence of categorical features, only methods that natively support mixed feature types are included in this comparison: CADEX, CCHVAE, SACE, and DiCE. Results are reported separately for MLP and Logistic Regression predictive models.

Tables~\ref{tab:cat_metrics_mlp} and~\ref{tab:cat_metrics_lr} report all metrics for MLP and Logistic Regression backbones, respectively, including coverage, validity, sparsity, probabilistic plausibility, log-density, LOF and Isolation Forest outlier scores, L2-Hamming proximity, and computation time.

Across all datasets and both predictive models, all methods achieve perfect coverage. Validity is near-perfect for most methods, with the notable exception of CADEX, which on Adult Census achieves only $\sim$0.51 validity (MLP), indicating that its constrained adversarial perturbations are often insufficient to cross the decision boundary in this setting. DiCE also exhibits occasional drops in validity, particularly under the Logistic Regression backbone (e.g., 0.73 on German Credit, 0.81 on Credit Default, and 0.79 on Lending Club).

In terms of proximity, CADEX tends to achieve the lowest L2-Hamming distances on several datasets (e.g., Adult Census, Bank Marketing, Give Me Some Credit), while DiCE achieves the smallest perturbations on German Credit. On Lending Club, CCHVAE achieves the best proximity across both backbones. However, low proximity does not always correspond to high plausibility: on datasets with large feature ranges such as Credit Default and Lending Club, CADEX and DiCE exhibit extremely low log-density values, indicating that their counterfactuals, while close in normalized space, may fall in low-density regions.

SACE stands out for its consistently strong plausibility scores. It achieves the highest probabilistic plausibility and log-density values across most datasets (e.g., 0.93 and $-109.47$ on Adult Census with MLP; 0.97 and $22.11$ on Give Me Some Credit), along with the lowest LOF scores, suggesting that its case-based retrieval strategy naturally produces counterfactuals close to the training data manifold. CCHVAE also achieves competitive plausibility on smaller datasets such as Law and German Credit (e.g., 0.92 on Law with MLP, 0.83 on German Credit with LR).

SACE and CCHVAE are also among the fastest methods, with SACE completing in under 1 second on smaller datasets (e.g., German Credit, Law). CADEX and DiCE require longer computation times, particularly on larger datasets such as Adult Census and Lending Club.

\begin{table*}[ht]
\centering
\tiny
\caption{Local methods on mixed-type datasets: metrics (MLP).}
\label{tab:cat_metrics_mlp}
\begin{center}
\begin{sc}
\begin{scriptsize}
\begin{tabular}{l|l|rrrrrrrrr}
\toprule
 & Method & Cov.$\uparrow$ & Valid.$\uparrow$ & Spars.$\uparrow$ & Prob. Plaus.$\uparrow$ & Log Dens.$\uparrow$ & LOF$\downarrow$ & IsoForest$\uparrow$ & L2-Ham.$\downarrow$ & Time(s)$\downarrow$ \\
\midrule
\multirow{5}{*}{Adult Census} & CADEX & $\boldsymbol{1.00\pm{0.00}}$ & $0.51\pm{0.02}$ & $0.01\pm{0.00}$ & $0.62\pm{0.04}$ & $-147.00\pm{36.30}$ & $\boldsymbol{2.41\pm{0.06}}$ & $0.13\pm{0.00}$ & $\boldsymbol{0.00\pm{0.00}}$ & $304.92\pm{110.68}$ \\
 & CCHVAE & $\boldsymbol{1.00\pm{0.00}}$ & $\boldsymbol{1.00\pm{0.00}}$ & $0.04\pm{0.00}$ & $0.77\pm{0.04}$ & $-214.42\pm{136.54}$ & $5.67\pm{0.90}$ & $\boldsymbol{0.14\pm{0.00}}$ & $0.02\pm{0.00}$ & $\boldsymbol{45.36\pm{3.86}}$ \\
 & SACE & $\boldsymbol{1.00\pm{0.00}}$ & $\boldsymbol{1.00\pm{0.00}}$ & $\boldsymbol{0.07\pm{0.00}}$ & $\boldsymbol{0.93\pm{0.04}}$ & $\boldsymbol{-109.47\pm{39.10}}$ & $3.17\pm{4.32}$ & $0.13\pm{0.01}$ & $0.06\pm{0.00}$ & $52.38\pm{2.27}$ \\
 & DiCE & $\boldsymbol{1.00\pm{0.00}}$ & $0.99\pm{0.01}$ & $0.01\pm{0.00}$ & $0.35\pm{0.06}$ & $-1004.90\pm{1272.17}$ & $8.28\pm{0.32}$ & $0.12\pm{0.00}$ & $0.01\pm{0.00}$ & $162.59\pm{9.31}$ \\
\midrule
\multirow{5}{*}{Bank Marketing} & CADEX & $\boldsymbol{1.00\pm{0.00}}$ & $\boldsymbol{1.00\pm{0.00}}$ & $0.14\pm{0.00}$ & $0.27\pm{0.04}$ & $-63.91\pm{4.73}$ & $1.88\pm{0.05}$ & $0.04\pm{0.00}$ & $\boldsymbol{0.02\pm{0.00}}$ & $9.22\pm{1.74}$ \\
 & CCHVAE & $\boldsymbol{1.00\pm{0.00}}$ & $\boldsymbol{1.00\pm{0.00}}$ & $0.16\pm{0.00}$ & $0.67\pm{0.07}$ & $-40.50\pm{2.03}$ & $2.48\pm{0.16}$ & $0.07\pm{0.01}$ & $0.07\pm{0.00}$ & $\boldsymbol{2.90\pm{0.68}}$ \\
 & SACE & $\boldsymbol{1.00\pm{0.00}}$ & $\boldsymbol{1.00\pm{0.00}}$ & $\boldsymbol{0.30\pm{0.00}}$ & $\boldsymbol{0.98\pm{0.01}}$ & $\boldsymbol{-27.52\pm{2.52}}$ & $\boldsymbol{1.11\pm{0.03}}$ & $\boldsymbol{0.09\pm{0.01}}$ & $0.26\pm{0.00}$ & $18.28\pm{2.77}$ \\
 & DiCE & $\boldsymbol{1.00\pm{0.00}}$ & $0.93\pm{0.09}$ & $0.04\pm{0.00}$ & $0.13\pm{0.02}$ & $-94.68\pm{11.55}$ & $3.42\pm{0.23}$ & $0.01\pm{0.00}$ & $0.06\pm{0.01}$ & $34.14\pm{7.28}$ \\
\midrule
\multirow{5}{*}{Credit Default} & CADEX & $\boldsymbol{1.00\pm{0.00}}$ & $\boldsymbol{1.00\pm{0.00}}$ & $0.16\pm{0.00}$ & $0.01\pm{0.00}$ & $-1457.68\pm{553.77}$ & $3.07\pm{0.11}$ & $-0.05\pm{0.01}$ & $\boldsymbol{0.06\pm{0.01}}$ & $34.53\pm{3.29}$ \\
 & CCHVAE & $\boldsymbol{1.00\pm{0.00}}$ & $\boldsymbol{1.00\pm{0.00}}$ & $0.18\pm{0.00}$ & $\boldsymbol{0.46\pm{0.06}}$ & $-413.44\pm{555.21}$ & $4.55\pm{0.23}$ & $0.05\pm{0.00}$ & $\boldsymbol{0.06\pm{0.00}}$ & $\boldsymbol{12.70\pm{2.15}}$ \\
 & SACE & $\boldsymbol{1.00\pm{0.00}}$ & $\boldsymbol{1.00\pm{0.00}}$ & $\boldsymbol{0.26\pm{0.01}}$ & $0.28\pm{0.41}$ & $\boldsymbol{39.91\pm{37.24}}$ & $\boldsymbol{1.05\pm{0.02}}$ & $\boldsymbol{0.08\pm{0.04}}$ & $0.16\pm{0.00}$ & $18.15\pm{0.64}$ \\
 & DiCE & $\boldsymbol{1.00\pm{0.00}}$ & $0.95\pm{0.08}$ & $0.02\pm{0.00}$ & $0.02\pm{0.01}$ & $-8987.03\pm{5645.83}$ & $4.22\pm{0.34}$ & $0.00\pm{0.00}$ & $0.10\pm{0.01}$ & $60.59\pm{2.93}$ \\
\midrule
\multirow{5}{*}{German Credit} & CADEX & $\boldsymbol{1.00\pm{0.00}}$ & $\boldsymbol{1.00\pm{0.00}}$ & $0.21\pm{0.01}$ & $0.35\pm{0.05}$ & $-73.58\pm{5.70}$ & $1.08\pm{0.00}$ & $0.01\pm{0.01}$ & $0.13\pm{0.01}$ & $4.51\pm{0.22}$ \\
 & CCHVAE & $\boldsymbol{1.00\pm{0.00}}$ & $\boldsymbol{1.00\pm{0.00}}$ & $0.28\pm{0.00}$ & $0.68\pm{0.40}$ & $-56.56\pm{16.10}$ & $\boldsymbol{0.98\pm{0.00}}$ & $\boldsymbol{0.12\pm{0.00}}$ & $0.24\pm{0.01}$ & $1.20\pm{1.39}$ \\
 & SACE & $\boldsymbol{1.00\pm{0.00}}$ & $\boldsymbol{1.00\pm{0.00}}$ & $\boldsymbol{0.32\pm{0.01}}$ & $\boldsymbol{0.93\pm{0.02}}$ & $\boldsymbol{-44.13\pm{6.43}}$ & $1.04\pm{0.01}$ & $0.04\pm{0.02}$ & $0.35\pm{0.02}$ & $\boldsymbol{0.03\pm{0.00}}$ \\
 & DiCE & $\boldsymbol{1.00\pm{0.00}}$ & $0.97\pm{0.03}$ & $0.08\pm{0.01}$ & $0.11\pm{0.06}$ & $-129.32\pm{26.54}$ & $1.12\pm{0.02}$ & $-0.01\pm{0.01}$ & $\boldsymbol{0.10\pm{0.02}}$ & $6.49\pm{0.99}$ \\
\midrule
\multirow{5}{*}{GMC} & CADEX & $\boldsymbol{1.00\pm{0.00}}$ & $0.98\pm{0.01}$ & $0.16\pm{0.01}$ & $0.27\pm{0.09}$ & $-703.55\pm{980.46}$ & $2.34\pm{0.86}$ & $-0.01\pm{0.03}$ & $\boldsymbol{0.02\pm{0.02}}$ & $35.20\pm{13.82}$ \\
 & CCHVAE & $\boldsymbol{1.00\pm{0.00}}$ & $\boldsymbol{1.00\pm{0.00}}$ & $0.20\pm{0.01}$ & $0.69\pm{0.12}$ & $-57.27\pm{55.47}$ & $7.72\pm{1.98}$ & $0.08\pm{0.01}$ & $0.08\pm{0.01}$ & $29.28\pm{3.58}$ \\
 & SACE & $\boldsymbol{1.00\pm{0.00}}$ & $\boldsymbol{1.00\pm{0.00}}$ & $\boldsymbol{0.23\pm{0.00}}$ & $\boldsymbol{0.97\pm{0.02}}$ & $\boldsymbol{22.11\pm{4.61}}$ & $\boldsymbol{1.42\pm{0.24}}$ & $\boldsymbol{0.10\pm{0.01}}$ & $0.12\pm{0.00}$ & $\boldsymbol{9.88\pm{0.47}}$ \\
 & DiCE & $\boldsymbol{1.00\pm{0.00}}$ & $0.98\pm{0.02}$ & $0.04\pm{0.00}$ & $0.06\pm{0.01}$ & $-1081.51\pm{1542.37}$ & $11.14\pm{0.99}$ & $-0.01\pm{0.01}$ & $0.10\pm{0.02}$ & $118.58\pm{31.94}$ \\
\midrule
\multirow{5}{*}{Law} & CADEX & $\boldsymbol{1.00\pm{0.00}}$ & $\boldsymbol{1.00\pm{0.00}}$ & $0.28\pm{0.00}$ & $0.66\pm{0.04}$ & $-12.76\pm{0.61}$ & $1.68\pm{0.21}$ & $0.05\pm{0.00}$ & $0.09\pm{0.01}$ & $7.07\pm{0.52}$ \\
 & CCHVAE & $\boldsymbol{1.00\pm{0.00}}$ & $\boldsymbol{1.00\pm{0.00}}$ & $0.24\pm{0.00}$ & $\boldsymbol{0.92\pm{0.02}}$ & $\boldsymbol{-10.16\pm{0.61}}$ & $1.22\pm{0.15}$ & $\boldsymbol{0.12\pm{0.00}}$ & $0.08\pm{0.01}$ & $4.26\pm{1.43}$ \\
 & SACE & $\boldsymbol{1.00\pm{0.00}}$ & $\boldsymbol{1.00\pm{0.00}}$ & $\boldsymbol{0.33\pm{0.08}}$ & $0.68\pm{0.30}$ & $-12.92\pm{2.39}$ & $1.83\pm{1.77}$ & $0.04\pm{0.07}$ & $0.24\pm{0.07}$ & $\boldsymbol{0.17\pm{0.00}}$ \\
 & DiCE & $\boldsymbol{1.00\pm{0.00}}$ & $0.98\pm{0.04}$ & $0.11\pm{0.00}$ & $0.33\pm{0.01}$ & $-17.46\pm{1.13}$ & $5.21\pm{0.50}$ & $0.00\pm{0.00}$ & $0.13\pm{0.00}$ & $4.79\pm{0.14}$ \\
\midrule
\multirow{5}{*}{Lending Club} & CADEX & $\boldsymbol{1.00\pm{0.00}}$ & $\boldsymbol{1.00\pm{0.00}}$ & $\boldsymbol{0.43\pm{0.01}}$ & $0.01\pm{0.01}$ & $-2.91\times 10^{8}\pm{6.51\times 10^{8}}$ & $3.25\pm{0.67}$ & $-0.01\pm{0.01}$ & $0.30\pm{0.01}$ & $270.51\pm{17.68}$ \\
 & CCHVAE & $\boldsymbol{1.00\pm{0.00}}$ & $\boldsymbol{1.00\pm{0.00}}$ & $0.33\pm{0.01}$ & $0.70\pm{0.11}$ & $-27.51\pm{12.79}$ & $1.87\pm{0.18}$ & $\boldsymbol{0.05\pm{0.00}}$ & $\boldsymbol{0.18\pm{0.01}}$ & $113.32\pm{9.60}$ \\
 & SACE & $\boldsymbol{1.00\pm{0.00}}$ & $\boldsymbol{1.00\pm{0.00}}$ & $0.42\pm{0.01}$ & $\boldsymbol{0.86\pm{0.11}}$ & $\boldsymbol{-13.83\pm{9.94}}$ & $\boldsymbol{1.06\pm{0.05}}$ & $\boldsymbol{0.05\pm{0.01}}$ & $0.32\pm{0.01}$ & $\boldsymbol{28.22\pm{0.48}}$ \\
 & DiCE & $\boldsymbol{1.00\pm{0.00}}$ & $0.94\pm{0.12}$ & $0.15\pm{0.02}$ & $0.04\pm{0.03}$ & $-3760.84\pm{6580.03}$ & $6.78\pm{0.69}$ & $-0.03\pm{0.01}$ & $0.24\pm{0.02}$ & $373.73\pm{64.16}$ \\
\bottomrule
\end{tabular}
\end{scriptsize}
\end{sc}
\end{center}
\end{table*}

\begin{table*}[ht]
\centering
\caption{Local methods on mixed-type datasets: metrics (LR).}
\label{tab:cat_metrics_lr}
\begin{center}
\begin{sc}
\begin{scriptsize}
\begin{tabular}{l|l|rrrrrrrrr}
\toprule
 & Method & Cov.$\uparrow$ & Valid.$\uparrow$ & Spars.$\uparrow$ & Prob. Plaus.$\uparrow$ & Log Dens.$\uparrow$ & LOF$\downarrow$ & IsoForest$\uparrow$ & L2-Ham.$\downarrow$ & Time(s)$\downarrow$ \\
\midrule
\multirow{5}{*}{Adult Census} & CADEX & $\boldsymbol{1.00\pm{0.00}}$ & $0.56\pm{0.02}$ & $0.01\pm{0.00}$ & $0.61\pm{0.04}$ & $-166.58\pm{7.96}$ & $\boldsymbol{2.37\pm{0.10}}$ & $0.13\pm{0.00}$ & $\boldsymbol{0.00\pm{0.00}}$ & $75.51\pm{3.50}$ \\
 & CCHVAE & $\boldsymbol{1.00\pm{0.00}}$ & $\boldsymbol{1.00\pm{0.00}}$ & $0.03\pm{0.00}$ & $0.75\pm{0.03}$ & $-142.85\pm{19.68}$ & $6.49\pm{0.98}$ & $\boldsymbol{0.14\pm{0.00}}$ & $0.02\pm{0.00}$ & $\boldsymbol{37.67\pm{2.21}}$ \\
 & SACE & $\boldsymbol{1.00\pm{0.00}}$ & $\boldsymbol{1.00\pm{0.00}}$ & $\boldsymbol{0.07\pm{0.00}}$ & $\boldsymbol{0.92\pm{0.05}}$ & $\boldsymbol{-118.19\pm{10.11}}$ & $3.10\pm{4.32}$ & $0.13\pm{0.01}$ & $0.06\pm{0.00}$ & $50.20\pm{0.65}$ \\
 & DiCE & $\boldsymbol{1.00\pm{0.00}}$ & $0.97\pm{0.04}$ & $0.01\pm{0.00}$ & $0.35\pm{0.04}$ & $-461.00\pm{145.25}$ & $8.03\pm{0.21}$ & $0.12\pm{0.00}$ & $0.01\pm{0.00}$ & $158.63\pm{3.98}$ \\
\midrule
\multirow{5}{*}{Bank Marketing} & CADEX & $\boldsymbol{1.00\pm{0.00}}$ & $\boldsymbol{1.00\pm{0.00}}$ & $0.15\pm{0.00}$ & $0.21\pm{0.04}$ & $-62.52\pm{9.35}$ & $1.66\pm{0.12}$ & $0.03\pm{0.00}$ & $\boldsymbol{0.02\pm{0.01}}$ & $4.17\pm{0.60}$ \\
 & CCHVAE & $\boldsymbol{1.00\pm{0.00}}$ & $\boldsymbol{1.00\pm{0.00}}$ & $0.16\pm{0.00}$ & $0.61\pm{0.06}$ & $-73.85\pm{70.46}$ & $2.24\pm{0.15}$ & $0.07\pm{0.00}$ & $0.07\pm{0.00}$ & $\boldsymbol{1.51\pm{0.11}}$ \\
 & SACE & $\boldsymbol{1.00\pm{0.00}}$ & $\boldsymbol{1.00\pm{0.00}}$ & $\boldsymbol{0.31\pm{0.00}}$ & $\boldsymbol{0.99\pm{0.01}}$ & $\boldsymbol{-23.93\pm{7.90}}$ & $\boldsymbol{1.10\pm{0.03}}$ & $\boldsymbol{0.09\pm{0.01}}$ & $0.27\pm{0.00}$ & $13.27\pm{0.66}$ \\
 & DiCE & $\boldsymbol{1.00\pm{0.00}}$ & $0.91\pm{0.08}$ & $0.04\pm{0.00}$ & $0.09\pm{0.02}$ & $-318.65\pm{502.93}$ & $2.88\pm{0.19}$ & $0.01\pm{0.00}$ & $0.08\pm{0.01}$ & $24.85\pm{1.25}$ \\
\midrule
\multirow{5}{*}{Credit Default} & CADEX & $\boldsymbol{1.00\pm{0.00}}$ & $\boldsymbol{1.00\pm{0.00}}$ & $0.16\pm{0.01}$ & $0.02\pm{0.01}$ & $-3376.12\pm{2761.91}$ & $2.45\pm{0.21}$ & $-0.05\pm{0.02}$ & $\boldsymbol{0.06\pm{0.02}}$ & $16.21\pm{3.99}$ \\
 & CCHVAE & $\boldsymbol{1.00\pm{0.00}}$ & $\boldsymbol{1.00\pm{0.00}}$ & $0.19\pm{0.00}$ & $\boldsymbol{0.32\pm{0.10}}$ & $-274.46\pm{144.35}$ & $4.63\pm{0.68}$ & $0.05\pm{0.01}$ & $\boldsymbol{0.06\pm{0.00}}$ & $\boldsymbol{8.99\pm{0.99}}$ \\
 & SACE & $\boldsymbol{1.00\pm{0.00}}$ & $\boldsymbol{1.00\pm{0.00}}$ & $\boldsymbol{0.26\pm{0.01}}$ & $0.20\pm{0.45}$ & $\boldsymbol{48.27\pm{26.09}}$ & $\boldsymbol{1.05\pm{0.02}}$ & $\boldsymbol{0.08\pm{0.04}}$ & $0.16\pm{0.00}$ & $18.07\pm{0.42}$ \\
 & DiCE & $\boldsymbol{1.00\pm{0.00}}$ & $0.81\pm{0.23}$ & $0.02\pm{0.00}$ & $0.02\pm{0.01}$ & $-1.23\times 10^{6}\pm{2.73\times 10^{6}}$ & $4.08\pm{0.29}$ & $0.00\pm{0.00}$ & $0.10\pm{0.01}$ & $58.96\pm{2.38}$ \\
\midrule
\multirow{5}{*}{German Credit} & CADEX & $\boldsymbol{1.00\pm{0.00}}$ & $\boldsymbol{1.00\pm{0.00}}$ & $\boldsymbol{0.35\pm{0.03}}$ & $0.19\pm{0.03}$ & $-90.66\pm{10.56}$ & $1.14\pm{0.02}$ & $-0.01\pm{0.01}$ & $0.28\pm{0.03}$ & $1.90\pm{0.09}$ \\
 & CCHVAE & $\boldsymbol{1.00\pm{0.00}}$ & $\boldsymbol{1.00\pm{0.00}}$ & $0.28\pm{0.00}$ & $0.83\pm{0.10}$ & $-48.90\pm{5.03}$ & $\boldsymbol{0.98\pm{0.01}}$ & $\boldsymbol{0.12\pm{0.00}}$ & $0.25\pm{0.01}$ & $0.46\pm{0.41}$ \\
 & SACE & $\boldsymbol{1.00\pm{0.00}}$ & $\boldsymbol{1.00\pm{0.00}}$ & $0.32\pm{0.01}$ & $\boldsymbol{0.92\pm{0.05}}$ & $\boldsymbol{-45.30\pm{6.75}}$ & $1.03\pm{0.01}$ & $0.04\pm{0.02}$ & $0.36\pm{0.01}$ & $\boldsymbol{0.03\pm{0.00}}$ \\
 & DiCE & $\boldsymbol{1.00\pm{0.00}}$ & $0.73\pm{0.21}$ & $0.08\pm{0.03}$ & $0.13\pm{0.08}$ & $-129.90\pm{39.08}$ & $1.13\pm{0.04}$ & $-0.01\pm{0.02}$ & $\boldsymbol{0.10\pm{0.04}}$ & $6.58\pm{2.47}$ \\
\midrule
\multirow{5}{*}{GMC} & CADEX & $\boldsymbol{1.00\pm{0.00}}$ & $0.79\pm{0.03}$ & $0.13\pm{0.01}$ & $0.21\pm{0.04}$ & $-1996.84\pm{3794.71}$ & $2.09\pm{0.11}$ & $0.02\pm{0.01}$ & $\boldsymbol{0.01\pm{0.00}}$ & $35.90\pm{3.58}$ \\
 & CCHVAE & $\boldsymbol{1.00\pm{0.00}}$ & $\boldsymbol{1.00\pm{0.00}}$ & $0.20\pm{0.00}$ & $0.47\pm{0.11}$ & $-257.76\pm{374.80}$ & $14.86\pm{3.79}$ & $0.07\pm{0.01}$ & $0.07\pm{0.01}$ & $16.74\pm{1.41}$ \\
 & SACE & $\boldsymbol{1.00\pm{0.00}}$ & $\boldsymbol{1.00\pm{0.00}}$ & $\boldsymbol{0.23\pm{0.00}}$ & $\boldsymbol{0.98\pm{0.01}}$ & $\boldsymbol{22.77\pm{11.02}}$ & $\boldsymbol{1.42\pm{0.24}}$ & $\boldsymbol{0.10\pm{0.01}}$ & $0.12\pm{0.00}$ & $\boldsymbol{9.80\pm{0.45}}$ \\
 & DiCE & $\boldsymbol{1.00\pm{0.00}}$ & $0.92\pm{0.06}$ & $0.03\pm{0.00}$ & $0.05\pm{0.01}$ & $-181533.88\pm{403595.91}$ & $11.07\pm{0.75}$ & $0.00\pm{0.00}$ & $0.08\pm{0.00}$ & $58.36\pm{2.48}$ \\
\midrule
\multirow{5}{*}{Law} & CADEX & $\boldsymbol{1.00\pm{0.00}}$ & $\boldsymbol{1.00\pm{0.00}}$ & $0.28\pm{0.01}$ & $0.68\pm{0.05}$ & $-12.77\pm{0.97}$ & $1.58\pm{0.15}$ & $0.06\pm{0.01}$ & $0.10\pm{0.01}$ & $3.90\pm{0.56}$ \\
 & CCHVAE & $\boldsymbol{1.00\pm{0.00}}$ & $\boldsymbol{1.00\pm{0.00}}$ & $0.24\pm{0.00}$ & $\boldsymbol{0.93\pm{0.03}}$ & $\boldsymbol{-10.52\pm{0.95}}$ & $\boldsymbol{1.02\pm{0.07}}$ & $\boldsymbol{0.11\pm{0.00}}$ & $0.08\pm{0.01}$ & $2.78\pm{0.53}$ \\
 & SACE & $\boldsymbol{1.00\pm{0.00}}$ & $\boldsymbol{1.00\pm{0.00}}$ & $\boldsymbol{0.33\pm{0.08}}$ & $0.61\pm{0.30}$ & $-13.74\pm{2.11}$ & $1.82\pm{1.77}$ & $0.04\pm{0.07}$ & $0.24\pm{0.07}$ & $\boldsymbol{0.16\pm{0.01}}$ \\
 & DiCE & $\boldsymbol{1.00\pm{0.00}}$ & $0.92\pm{0.04}$ & $0.11\pm{0.00}$ & $0.28\pm{0.03}$ & $-17.70\pm{1.05}$ & $5.38\pm{0.27}$ & $0.00\pm{0.01}$ & $0.13\pm{0.00}$ & $4.92\pm{0.19}$ \\
\midrule
\multirow{5}{*}{Lending Club} & CADEX & $\boldsymbol{1.00\pm{0.00}}$ & $\boldsymbol{1.00\pm{0.00}}$ & $0.37\pm{0.04}$ & $0.03\pm{0.00}$ & $-1.90\times 10^{8}\pm{4.28\times 10^{8}}$ & $3.10\pm{1.02}$ & $-0.01\pm{0.02}$ & $0.25\pm{0.04}$ & $90.70\pm{3.01}$ \\
 & CCHVAE & $\boldsymbol{1.00\pm{0.00}}$ & $\boldsymbol{1.00\pm{0.00}}$ & $0.34\pm{0.00}$ & $\boldsymbol{0.76\pm{0.15}}$ & $-6.56\times 10^{6}\pm{1.48\times 10^{7}}$ & $2.33\pm{0.32}$ & $\boldsymbol{0.05\pm{0.00}}$ & $\boldsymbol{0.19\pm{0.01}}$ & $69.45\pm{2.18}$ \\
 & SACE & $\boldsymbol{1.00\pm{0.00}}$ & $\boldsymbol{1.00\pm{0.00}}$ & $\boldsymbol{0.42\pm{0.02}}$ & $0.71\pm{0.13}$ & $\boldsymbol{-16.38\pm{8.75}}$ & $\boldsymbol{1.08\pm{0.01}}$ & $\boldsymbol{0.05\pm{0.02}}$ & $0.33\pm{0.04}$ & $\boldsymbol{27.79\pm{0.60}}$ \\
 & DiCE & $\boldsymbol{1.00\pm{0.00}}$ & $0.79\pm{0.22}$ & $0.08\pm{0.03}$ & $0.02\pm{0.01}$ & $-2.51\times 10^{10}\pm{5.66\times 10^{10}}$ & $6.29\pm{0.43}$ & $0.01\pm{0.01}$ & $0.21\pm{0.01}$ & $268.12\pm{99.87}$ \\
\bottomrule
\end{tabular}
\end{scriptsize}
\end{sc}
\end{center}
\end{table*}

\subsection{Local Methods on Numerical Datasets}

This subsection reports results for local counterfactual methods evaluated on six classification datasets with exclusively numerical features: Blobs, Digits, Moons, Wine, Audit, and HELOC. Since these datasets do not contain categorical features, additional methods that operate only on continuous inputs are included: Artelt, CEGP, CEM, and Wachter, alongside CADEX, CCHVAE, SACE, DiCE, and PPCEF. Results are reported separately for MLP and Logistic Regression predictive models. Tables~\ref{tab:num_metrics_mlp} and~\ref{tab:num_metrics_lr} report all metrics for MLP and Logistic Regression backbones, respectively.

All methods achieve perfect coverage except Artelt, which returns zero coverage on several datasets (Blobs, Moons, HELOC with MLP), indicating a failure to produce counterfactual candidates in these settings. CEM also fails to produce results on Audit with MLP. Validity is generally high across most methods, though DiCE shows reduced validity on some datasets (e.g., 0.80 on Wine, 0.41 on Audit with MLP). CEM also struggles with validity on Audit under Logistic Regression (0.02).

In terms of proximity, CADEX and CEGP consistently achieve the lowest L2 distances across most datasets. On Blobs, both CADEX and CEGP achieve the smallest perturbations (0.34 with MLP). On HELOC, CADEX and CEGP achieve the lowest L2 distances (0.14 and 0.15 with MLP, respectively), while CEGP achieves the best proximity under LR (0.11). Wachter achieves competitive proximity, particularly on low-dimensional datasets such as Moons and Blobs, where it closely matches CADEX.

Plausibility metrics reveal strong differentiation. PPCEF achieves perfect or near-perfect probabilistic plausibility on low-dimensional datasets (1.00 on Blobs and Wine under both backbones; 1.00 on Moons with MLP, 0.90 with LR), along with the highest log-density values, demonstrating that its flow-based objective effectively constrains counterfactuals to high-density regions when the data manifold is well-captured. On HELOC, both PPCEF and CCHVAE achieve high plausibility (PPCEF: 1.00, CCHVAE: 0.99 with MLP). SACE similarly achieves strong plausibility on Digits and Audit, with the highest log-density and lowest LOF scores in these settings (e.g., 1.00 probabilistic plausibility and 60.70 log-density on Audit with MLP). In contrast, methods without explicit density objectives (Wachter, CADEX, CEGP) generally achieve low probabilistic plausibility, indicating that proximity-focused optimization alone does not ensure distributional alignment.

The Audit dataset exposes numerical instabilities: several methods exhibit extremely high LOF values (on the order of $10^{7}$), suggesting that counterfactuals fall far outside the training data manifold on this high-dimensional dataset with heterogeneous feature scales.

Wachter is consistently the fastest method, often completing in under 0.1 seconds, while CEGP and CEM require substantially longer computation times, particularly on larger datasets such as HELOC ($>$1,500s for CEGP).

\begin{table*}[ht]
\centering
\caption{Local methods on numerical datasets: metrics (MLP).}
\label{tab:num_metrics_mlp}
\begin{center}
\begin{sc}
\begin{scriptsize}
\begin{tabular}{l|l|rrrrrrrrr}
\toprule
 & Method & Cov.$\uparrow$ & Valid.$\uparrow$ & Sparse.$\uparrow$ & Prob. Plaus.$\uparrow$ & Log Dens.$\uparrow$ & LOF$\downarrow$ & IsoForest$\uparrow$ & L2$\downarrow$ & Time(s)$\downarrow$ \\
\midrule
\multirow{9}{*}{Blobs} & CADEX & $\boldsymbol{1.00\pm{0.00}}$ & $\boldsymbol{1.00\pm{0.00}}$ & $\boldsymbol{1.00\pm{0.00}}$ & $0.00\pm{0.00}$ & $-2.05\pm{0.51}$ & $1.64\pm{0.07}$ & $-0.06\pm{0.01}$ & $\boldsymbol{0.34\pm{0.03}}$ & $5.13\pm{0.40}$ \\
 & CCHVAE & $\boldsymbol{1.00\pm{0.00}}$ & $\boldsymbol{1.00\pm{0.00}}$ & $\boldsymbol{1.00\pm{0.00}}$ & $0.00\pm{0.00}$ & $-1.42\pm{0.46}$ & $1.55\pm{0.08}$ & $-0.05\pm{0.01}$ & $0.36\pm{0.03}$ & $1.54\pm{0.53}$ \\
 & SACE & $\boldsymbol{1.00\pm{0.00}}$ & $\boldsymbol{1.00\pm{0.00}}$ & $\boldsymbol{1.00\pm{0.00}}$ & $0.20\pm{0.45}$ & $1.23\pm{0.83}$ & $1.16\pm{0.12}$ & $0.01\pm{0.03}$ & $0.48\pm{0.09}$ & $\boldsymbol{0.05\pm{0.01}}$ \\
 & DiCE & $\boldsymbol{1.00\pm{0.00}}$ & $0.98\pm{0.03}$ & $0.76\pm{0.03}$ & $0.08\pm{0.04}$ & $-8.70\pm{1.69}$ & $3.13\pm{0.22}$ & $-0.11\pm{0.01}$ & $0.65\pm{0.01}$ & $2.06\pm{0.02}$ \\
 & PPCEF & $\boldsymbol{1.00\pm{0.00}}$ & $\boldsymbol{1.00\pm{0.00}}$ & $\boldsymbol{1.00\pm{0.00}}$ & $\boldsymbol{1.00\pm{0.00}}$ & $\boldsymbol{1.60\pm{0.09}}$ & $\boldsymbol{1.08\pm{0.02}}$ & $0.02\pm{0.01}$ & $0.48\pm{0.03}$ & $12.03\pm{0.12}$ \\
 & Artelt & $0.00\pm{0.00}$ & $\boldsymbol{1.00\pm{0.00}}$ & $\boldsymbol{1.00\pm{0.00}}$ & $0.00\pm{0.00}$ & -- & -- & $\boldsymbol{0.12\pm{0.00}}$ & -- & $21.47\pm{0.32}$ \\
 & CEGP & $\boldsymbol{1.00\pm{0.00}}$ & $0.99\pm{0.01}$ & $0.99\pm{0.01}$ & $0.00\pm{0.00}$ & $-2.36\pm{1.04}$ & $1.60\pm{0.07}$ & $-0.05\pm{0.00}$ & $\boldsymbol{0.34\pm{0.03}}$ & $176.78\pm{3.21}$ \\
 & CEM & $\boldsymbol{1.00\pm{0.00}}$ & $0.89\pm{0.05}$ & $0.49\pm{0.03}$ & $0.03\pm{0.03}$ & $-15.56\pm{2.70}$ & $3.40\pm{0.25}$ & $-0.10\pm{0.01}$ & $0.52\pm{0.04}$ & $78.92\pm{1.28}$ \\
 & WACH & $\boldsymbol{1.00\pm{0.00}}$ & $\boldsymbol{1.00\pm{0.00}}$ & $\boldsymbol{1.00\pm{0.00}}$ & $0.00\pm{0.00}$ & $-1.50\pm{0.47}$ & $1.60\pm{0.06}$ & $-0.05\pm{0.00}$ & $0.36\pm{0.03}$ & $0.11\pm{0.01}$ \\
\midrule
\multirow{9}{*}{Digits} & CADEX & $\boldsymbol{1.00\pm{0.00}}$ & $\boldsymbol{1.00\pm{0.00}}$ & $\boldsymbol{1.00\pm{0.00}}$ & $0.00\pm{0.00}$ & $-2416.38\pm{1886.15}$ & $1.33\pm{0.04}$ & $-0.04\pm{0.01}$ & $\boldsymbol{1.53\pm{0.12}}$ & $1.68\pm{0.13}$ \\
 & CCHVAE & $\boldsymbol{1.00\pm{0.00}}$ & $\boldsymbol{1.00\pm{0.00}}$ & $\boldsymbol{1.00\pm{0.00}}$ & $0.00\pm{0.00}$ & $-172.46\pm{93.75}$ & $1.26\pm{0.09}$ & $0.01\pm{0.02}$ & $2.38\pm{0.08}$ & $0.98\pm{1.01}$ \\
 & SACE & $\boldsymbol{1.00\pm{0.00}}$ & $\boldsymbol{1.00\pm{0.00}}$ & $0.64\pm{0.01}$ & $0.80\pm{0.45}$ & $\boldsymbol{76.98\pm{15.04}}$ & $\boldsymbol{1.01\pm{0.05}}$ & $0.06\pm{0.02}$ & $3.65\pm{0.12}$ & $\boldsymbol{0.02\pm{0.01}}$ \\
 & DiCE & $\boldsymbol{1.00\pm{0.00}}$ & $0.93\pm{0.06}$ & $0.39\pm{0.02}$ & $0.00\pm{0.00}$ & $-3785.25\pm{3422.09}$ & $1.98\pm{0.07}$ & $-0.10\pm{0.00}$ & $2.96\pm{0.15}$ & $18.37\pm{1.23}$ \\
 & PPCEF & $\boldsymbol{1.00\pm{0.00}}$ & $\boldsymbol{1.00\pm{0.00}}$ & $\boldsymbol{1.00\pm{0.00}}$ & $\boldsymbol{0.98\pm{0.04}}$ & $71.55\pm{23.84}$ & $1.30\pm{0.08}$ & $0.00\pm{0.02}$ & $2.63\pm{0.29}$ & $14.40\pm{0.15}$ \\
 & Artelt & -- & $0.00\pm0.00$ & -- & -- & -- & -- & -- & -- & $2.06\pm0.02$ \\
 & CEGP & $\boldsymbol{1.00\pm{0.00}}$ & $0.99\pm{0.01}$ & $0.54\pm{0.03}$ & $0.00\pm{0.00}$ & $-714.25\pm{283.68}$ & $1.41\pm{0.03}$ & $-0.04\pm{0.01}$ & $1.76\pm{0.15}$ & $72.31\pm{1.72}$ \\
 & CEM & $\boldsymbol{1.00\pm{0.00}}$ & $0.96\pm{0.06}$ & $0.33\pm{0.01}$ & $0.00\pm{0.00}$ & $-6574.15\pm{4702.41}$ & $2.11\pm{0.07}$ & $-0.13\pm{0.01}$ & $3.17\pm{0.17}$ & $30.31\pm{0.63}$ \\
 & WACH & $\boldsymbol{1.00\pm{0.00}}$ & $\boldsymbol{1.00\pm{0.00}}$ & $\boldsymbol{1.00\pm{0.00}}$ & $0.00\pm{0.00}$ & $-2689.61\pm{2264.06}$ & $1.36\pm{0.04}$ & $-0.04\pm{0.01}$ & $1.60\pm{0.11}$ & $0.09\pm{0.01}$ \\
\midrule
\multirow{9}{*}{Moons} & CADEX & $\boldsymbol{1.00\pm{0.00}}$ & $\boldsymbol{1.00\pm{0.00}}$ & $\boldsymbol{1.00\pm{0.00}}$ & $0.00\pm{0.00}$ & $-5.30\pm{4.24}$ & $1.43\pm{0.05}$ & $0.00\pm{0.00}$ & $\boldsymbol{0.16\pm{0.02}}$ & $3.18\pm{0.37}$ \\
 & CCHVAE & $\boldsymbol{1.00\pm{0.00}}$ & $\boldsymbol{1.00\pm{0.00}}$ & $\boldsymbol{1.00\pm{0.00}}$ & $0.00\pm{0.00}$ & $-1.32\pm{1.12}$ & $1.37\pm{0.14}$ & $0.03\pm{0.01}$ & $0.29\pm{0.00}$ & $0.19\pm{0.20}$ \\
 & SACE & $\boldsymbol{1.00\pm{0.00}}$ & $\boldsymbol{1.00\pm{0.00}}$ & $\boldsymbol{1.00\pm{0.00}}$ & $0.80\pm{0.45}$ & $1.48\pm{0.27}$ & $1.02\pm{0.06}$ & $0.02\pm{0.02}$ & $0.31\pm{0.08}$ & $\boldsymbol{0.04\pm{0.00}}$ \\
 & DiCE & $\boldsymbol{1.00\pm{0.00}}$ & $\boldsymbol{1.00\pm{0.00}}$ & $0.72\pm{0.04}$ & $0.21\pm{0.05}$ & $-36.41\pm{57.71}$ & $1.72\pm{0.09}$ & $-0.05\pm{0.01}$ & $0.53\pm{0.03}$ & $1.89\pm{0.05}$ \\
 & PPCEF & $\boldsymbol{1.00\pm{0.00}}$ & $\boldsymbol{1.00\pm{0.00}}$ & $\boldsymbol{1.00\pm{0.00}}$ & $\boldsymbol{1.00\pm{0.00}}$ & $\boldsymbol{1.65\pm{0.06}}$ & $\boldsymbol{1.00\pm{0.01}}$ & $0.03\pm{0.00}$ & $0.27\pm{0.01}$ & $41.78\pm{0.47}$ \\
 & Artelt & $0.00\pm{0.00}$ & $\boldsymbol{1.00\pm{0.00}}$ & $\boldsymbol{1.00\pm{0.00}}$ & $0.00\pm{0.00}$ & -- & -- & $\boldsymbol{0.10\pm{0.00}}$ & -- & $51.01\pm{8.52}$ \\
 & CEGP & $\boldsymbol{1.00\pm{0.00}}$ & $0.99\pm{0.01}$ & $0.91\pm{0.03}$ & $0.01\pm{0.01}$ & $-4.53\pm{3.36}$ & $1.40\pm{0.06}$ & $0.00\pm{0.01}$ & $0.20\pm{0.05}$ & $169.75\pm{1.73}$ \\
 & CEM & $\boldsymbol{1.00\pm{0.00}}$ & $0.96\pm{0.03}$ & $0.57\pm{0.06}$ & $0.06\pm{0.05}$ & $-130.78\pm{223.07}$ & $2.19\pm{0.12}$ & $-0.08\pm{0.00}$ & $0.47\pm{0.10}$ & $74.77\pm{0.88}$ \\
 & WACH & $\boldsymbol{1.00\pm{0.00}}$ & $\boldsymbol{1.00\pm{0.00}}$ & $\boldsymbol{1.00\pm{0.00}}$ & $0.00\pm{0.00}$ & $-4.37\pm{3.85}$ & $1.41\pm{0.06}$ & $0.00\pm{0.00}$ & $0.18\pm{0.02}$ & $0.10\pm{0.01}$ \\
\midrule
\multirow{9}{*}{Wine} & CADEX & $\boldsymbol{1.00\pm{0.00}}$ & $\boldsymbol{1.00\pm{0.00}}$ & $\boldsymbol{1.00\pm{0.00}}$ & $0.02\pm{0.04}$ & $-3.96\pm{3.21}$ & $1.08\pm{0.04}$ & $0.05\pm{0.01}$ & $0.42\pm{0.02}$ & $0.29\pm{0.02}$ \\
 & CCHVAE & $\boldsymbol{1.00\pm{0.00}}$ & $\boldsymbol{1.00\pm{0.00}}$ & $\boldsymbol{1.00\pm{0.00}}$ & $0.41\pm{0.55}$ & $3.29\pm{3.27}$ & $1.01\pm{0.02}$ & $0.07\pm{0.01}$ & $0.85\pm{0.04}$ & $\boldsymbol{0.01\pm{0.00}}$ \\
 & SACE & $\boldsymbol{1.00\pm{0.00}}$ & $\boldsymbol{1.00\pm{0.00}}$ & $0.98\pm{0.02}$ & $0.00\pm{0.00}$ & $0.32\pm{3.35}$ & $1.27\pm{0.09}$ & $-0.03\pm{0.03}$ & $1.30\pm{0.11}$ & $\boldsymbol{0.01\pm{0.01}}$ \\
 & DiCE & $\boldsymbol{1.00\pm{0.00}}$ & $0.80\pm{0.20}$ & $0.19\pm{0.04}$ & $0.00\pm{0.00}$ & $-31.68\pm{25.40}$ & $1.32\pm{0.03}$ & $0.00\pm{0.01}$ & $0.88\pm{0.03}$ & $0.41\pm{0.04}$ \\
 & PPCEF & $\boldsymbol{1.00\pm{0.00}}$ & $\boldsymbol{1.00\pm{0.00}}$ & $\boldsymbol{1.00\pm{0.00}}$ & $\boldsymbol{1.00\pm{0.00}}$ & $\boldsymbol{5.66\pm{0.80}}$ & $\boldsymbol{1.00\pm{0.01}}$ & $\boldsymbol{0.08\pm{0.01}}$ & $0.56\pm{0.07}$ & $8.76\pm{0.09}$ \\
 & Artelt & $0.80\pm{0.45}$ & $0.82\pm{0.40}$ & $\boldsymbol{1.00\pm{0.00}}$ & $0.00\pm{0.00}$ & $-103.56\pm{116.65}$ & $2.12\pm{0.20}$ & $-0.04\pm{0.11}$ & $1.93\pm{0.16}$ & $2.29\pm{0.74}$ \\
 & CEGP & $\boldsymbol{1.00\pm{0.00}}$ & $\boldsymbol{1.00\pm{0.00}}$ & $0.73\pm{0.05}$ & $0.04\pm{0.05}$ & $-5.13\pm{5.91}$ & $1.06\pm{0.03}$ & $0.05\pm{0.01}$ & $\boldsymbol{0.41\pm{0.02}}$ & $17.99\pm{0.68}$ \\
 & CEM & $\boldsymbol{1.00\pm{0.00}}$ & $\boldsymbol{1.00\pm{0.00}}$ & $0.29\pm{0.03}$ & $0.00\pm{0.00}$ & $-30.80\pm{21.60}$ & $1.35\pm{0.06}$ & $-0.02\pm{0.02}$ & $0.68\pm{0.05}$ & $7.74\pm{0.26}$ \\
 & WACH & $\boldsymbol{1.00\pm{0.00}}$ & $\boldsymbol{1.00\pm{0.00}}$ & $\boldsymbol{1.00\pm{0.00}}$ & $0.04\pm{0.05}$ & $-3.01\pm{2.83}$ & $1.08\pm{0.05}$ & $0.05\pm{0.01}$ & $0.47\pm{0.02}$ & $0.05\pm{0.01}$ \\
\midrule
\multirow{9}{*}{Audit} & CADEX & $\boldsymbol{1.00\pm{0.00}}$ & $\boldsymbol{1.00\pm{0.01}}$ & $\boldsymbol{1.00\pm{0.01}}$ & $0.02\pm{0.04}$ & $-210452.76\pm{445022.59}$ & $6.22\times 10^{7}\pm{1.35\times 10^{8}}$ & $0.07\pm{0.01}$ & $\boldsymbol{0.65\pm{0.05}}$ & $1.89\pm{0.16}$ \\
 & CCHVAE & $\boldsymbol{1.00\pm{0.00}}$ & $\boldsymbol{1.00\pm{0.00}}$ & $\boldsymbol{1.00\pm{0.00}}$ & $0.00\pm{0.00}$ & $33.20\pm{8.31}$ & $86.57\pm{20.15}$ & $0.01\pm{0.01}$ & $1.32\pm{0.07}$ & $1.88\pm{0.18}$ \\
 & SACE & $\boldsymbol{1.00\pm{0.00}}$ & $\boldsymbol{1.00\pm{0.00}}$ & $0.62\pm{0.01}$ & $\boldsymbol{1.00\pm{0.00}}$ & $\boldsymbol{60.70\pm{3.47}}$ & $\boldsymbol{1.54\pm{1.09}}$ & $\boldsymbol{0.11\pm{0.01}}$ & $1.56\pm{0.12}$ & $\boldsymbol{0.03\pm{0.00}}$ \\
 & DiCE & $\boldsymbol{1.00\pm{0.00}}$ & $0.41\pm{0.06}$ & $0.03\pm{0.01}$ & $0.02\pm{0.02}$ & $-551.01\pm{655.15}$ & $6.24\times 10^{6}\pm{1.18\times 10^{7}}$ & $0.01\pm{0.01}$ & $0.82\pm{0.08}$ & $10.17\pm{0.87}$ \\
 & PPCEF & $\boldsymbol{1.00\pm{0.00}}$ & $0.99\pm{0.01}$ & $\boldsymbol{1.00\pm{0.00}}$ & $0.99\pm{0.01}$ & $51.67\pm{3.40}$ & $4.66\times 10^{6}\pm{1.02\times 10^{7}}$ & $0.07\pm{0.01}$ & $1.04\pm{0.14}$ & $42.96\pm{0.63}$ \\
 & Artelt & $\boldsymbol{1.00\pm{0.00}}$ & $0.12\pm{0.11}$ & $\boldsymbol{1.00\pm{0.00}}$ & $0.00\pm{0.00}$ & $-1447.93\pm{1000.19}$ & $33.32\pm{20.87}$ & $-0.05\pm{0.04}$ & $1.15\pm{0.33}$ & $170.78\pm{104.20}$ \\
 & CEGP & $\boldsymbol{1.00\pm{0.00}}$ & $0.78\pm{0.10}$ & $0.31\pm{0.05}$ & $0.12\pm{0.13}$ & $-472.46\pm{650.90}$ & $1.83\times 10^{7}\pm{3.66\times 10^{7}}$ & $0.02\pm{0.01}$ & $0.75\pm{0.09}$ & $111.21\pm{8.29}$ \\
 & CEM & -- & $0.00\pm{0.00}$ & -- & -- & -- & -- & -- & -- & $50.87\pm{3.13}$ \\
 & WACH & $\boldsymbol{1.00\pm{0.00}}$ & $\boldsymbol{1.00\pm{0.00}}$ & $\boldsymbol{1.00\pm{0.00}}$ & $0.02\pm{0.04}$ & $-9.65\times 10^{6}\pm{2.32\times 10^{7}}$ & $6.19\times 10^{7}\pm{1.35\times 10^{8}}$ & $0.07\pm{0.01}$ & $0.74\pm{0.04}$ & $0.23\pm{0.27}$ \\
\midrule
\multirow{9}{*}{HELOC} & CADEX & $\boldsymbol{1.00\pm{0.00}}$ & $\boldsymbol{1.00\pm{0.00}}$ & $\boldsymbol{1.00\pm{0.00}}$ & $0.29\pm{0.02}$ & $27.37\pm{1.09}$ & $1.16\pm{0.02}$ & $0.05\pm{0.01}$ & $\boldsymbol{0.14\pm{0.01}}$ & $6.85\pm{1.00}$ \\
 & CCHVAE & $\boldsymbol{1.00\pm{0.00}}$ & $\boldsymbol{1.00\pm{0.00}}$ & $\boldsymbol{1.00\pm{0.00}}$ & $0.99\pm{0.01}$ & $\boldsymbol{39.86\pm{1.68}}$ & $1.78\times 10^{6}\pm{3.86\times 10^{6}}$ & $0.12\pm{0.00}$ & $0.50\pm{0.02}$ & $6.82\pm{3.40}$ \\
 & SACE & $\boldsymbol{1.00\pm{0.00}}$ & $\boldsymbol{1.00\pm{0.00}}$ & $0.82\pm{0.04}$ & $0.22\pm{0.45}$ & $36.28\pm{6.41}$ & $\boldsymbol{1.04\pm{0.04}}$ & $0.07\pm{0.01}$ & $0.70\pm{0.03}$ & $3.30\pm{0.05}$ \\
 & DiCE & $\boldsymbol{1.00\pm{0.00}}$ & $0.94\pm{0.04}$ & $0.07\pm{0.00}$ & $0.01\pm{0.01}$ & $-158.61\pm{180.70}$ & $2.12\pm{0.08}$ & $0.04\pm{0.00}$ & $0.68\pm{0.01}$ & $43.08\pm{3.31}$ \\
 & PPCEF & $\boldsymbol{1.00\pm{0.00}}$ & $\boldsymbol{1.00\pm{0.00}}$ & $\boldsymbol{1.00\pm{0.00}}$ & $\boldsymbol{1.00\pm{0.00}}$ & $38.90\pm{0.80}$ & $1.09\pm{0.01}$ & $0.08\pm{0.00}$ & $0.24\pm{0.01}$ & $142.72\pm{3.65}$ \\
 & Artelt & $0.00\pm{0.00}$ & $\boldsymbol{1.00\pm{0.00}}$ & $\boldsymbol{1.00\pm{0.00}}$ & $0.00\pm{0.00}$ & -- & -- & $\boldsymbol{0.15\pm{0.00}}$ & -- & $413.63\pm{15.93}$ \\
 & CEGP & $\boldsymbol{1.00\pm{0.00}}$ & $\boldsymbol{1.00\pm{0.00}}$ & $0.46\pm{0.03}$ & $0.39\pm{0.05}$ & $29.79\pm{2.94}$ & $1.16\pm{0.02}$ & $0.06\pm{0.00}$ & $0.15\pm{0.01}$ & $1832.85\pm{132.87}$ \\
 & CEM & $\boldsymbol{1.00\pm{0.00}}$ & $\boldsymbol{1.00\pm{0.00}}$ & $0.24\pm{0.02}$ & $0.13\pm{0.02}$ & $4.41\pm{11.00}$ & $1.40\pm{0.05}$ & $0.03\pm{0.01}$ & $0.27\pm{0.02}$ & $743.68\pm{53.59}$ \\
 & WACH & $\boldsymbol{1.00\pm{0.00}}$ & $\boldsymbol{1.00\pm{0.00}}$ & $\boldsymbol{1.00\pm{0.00}}$ & $0.17\pm{0.02}$ & $22.42\pm{1.42}$ & $1.20\pm{0.02}$ & $0.04\pm{0.01}$ & $0.19\pm{0.01}$ & $\boldsymbol{0.08\pm{0.01}}$ \\
\bottomrule
\end{tabular}
\end{scriptsize}
\end{sc}
\end{center}
\end{table*}

\begin{table*}[ht]
\centering
\caption{Local methods on numerical datasets: metrics (LR).}
\label{tab:num_metrics_lr}
\begin{center}
\begin{sc}
\begin{scriptsize}
\begin{tabular}{l|l|rrrrrrrrr}
\toprule
 & Method & Cov.$\uparrow$ & Valid.$\uparrow$ & Sparse.$\uparrow$ & Prob. Plaus.$\uparrow$ & Log Dens.$\uparrow$ & LOF$\downarrow$ & IsoForest$\uparrow$ & L2$\downarrow$ & Time(s)$\downarrow$ \\
\midrule
\multirow{9}{*}{Blobs} & CADEX & $\boldsymbol{1.00\pm{0.00}}$ & $\boldsymbol{1.00\pm{0.00}}$ & $\boldsymbol{1.00\pm{0.00}}$ & $0.00\pm{0.00}$ & $-2.75\pm{0.32}$ & $1.65\pm{0.07}$ & $-0.06\pm{0.01}$ & $\boldsymbol{0.32\pm{0.02}}$ & $1.96\pm{0.12}$ \\
 & CCHVAE & $\boldsymbol{1.00\pm{0.00}}$ & $\boldsymbol{1.00\pm{0.00}}$ & $\boldsymbol{1.00\pm{0.00}}$ & $0.00\pm{0.00}$ & $-1.98\pm{0.17}$ & $1.58\pm{0.11}$ & $-0.05\pm{0.01}$ & $0.34\pm{0.02}$ & $0.64\pm{0.28}$ \\
 & SACE & $\boldsymbol{1.00\pm{0.00}}$ & $\boldsymbol{1.00\pm{0.00}}$ & $\boldsymbol{1.00\pm{0.00}}$ & $0.20\pm{0.45}$ & $1.28\pm{0.76}$ & $1.16\pm{0.12}$ & $0.01\pm{0.03}$ & $0.48\pm{0.09}$ & $\boldsymbol{0.04\pm{0.00}}$ \\
 & DiCE & $\boldsymbol{1.00\pm{0.00}}$ & $0.99\pm{0.03}$ & $0.75\pm{0.03}$ & $0.08\pm{0.03}$ & $-9.81\pm{4.99}$ & $3.02\pm{0.21}$ & $-0.11\pm{0.01}$ & $0.64\pm{0.03}$ & $1.99\pm{0.02}$ \\
 & PPCEF & $\boldsymbol{1.00\pm{0.00}}$ & $\boldsymbol{1.00\pm{0.00}}$ & $\boldsymbol{1.00\pm{0.00}}$ & $\boldsymbol{1.00\pm{0.00}}$ & $\boldsymbol{1.62\pm{0.08}}$ & $\boldsymbol{1.09\pm{0.02}}$ & $\boldsymbol{0.02\pm{0.01}}$ & $0.48\pm{0.03}$ & $11.17\pm{0.25}$ \\
 & Artelt & $\boldsymbol{1.00\pm{0.00}}$ & $\boldsymbol{1.00\pm{0.00}}$ & $\boldsymbol{1.00\pm{0.00}}$ & $0.00\pm{0.00}$ & $-7.25\pm{1.28}$ & $2.49\pm{0.29}$ & $-0.08\pm{0.01}$ & $0.40\pm{0.01}$ & $1.21\pm{0.12}$ \\
 & CEGP & $\boldsymbol{1.00\pm{0.00}}$ & $\boldsymbol{1.00\pm{0.00}}$ & $\boldsymbol{1.00\pm{0.00}}$ & $0.00\pm{0.00}$ & $-2.73\pm{0.54}$ & $1.62\pm{0.07}$ & $-0.06\pm{0.01}$ & $\boldsymbol{0.32\pm{0.02}}$ & $143.12\pm{3.34}$ \\
 & CEM & $\boldsymbol{1.00\pm{0.00}}$ & $0.96\pm{0.02}$ & $0.54\pm{0.02}$ & $0.04\pm{0.03}$ & $-16.11\pm{3.40}$ & $3.43\pm{0.18}$ & $-0.11\pm{0.01}$ & $0.51\pm{0.02}$ & $63.78\pm{0.59}$ \\
 & WACH & $\boldsymbol{1.00\pm{0.00}}$ & $\boldsymbol{1.00\pm{0.00}}$ & $\boldsymbol{1.00\pm{0.00}}$ & $0.00\pm{0.00}$ & $-1.81\pm{0.81}$ & $1.63\pm{0.08}$ & $-0.06\pm{0.01}$ & $0.34\pm{0.02}$ & $0.11\pm{0.02}$ \\
\midrule
\multirow{9}{*}{Digits} & CADEX & $\boldsymbol{1.00\pm{0.00}}$ & $\boldsymbol{1.00\pm{0.00}}$ & $\boldsymbol{1.00\pm{0.00}}$ & $0.00\pm{0.00}$ & $-1680.83\pm{982.76}$ & $1.25\pm{0.04}$ & $-0.04\pm{0.01}$ & $\boldsymbol{1.26\pm{0.06}}$ & $0.45\pm{0.02}$ \\
 & CCHVAE & $\boldsymbol{1.00\pm{0.00}}$ & $\boldsymbol{1.00\pm{0.00}}$ & $\boldsymbol{1.00\pm{0.00}}$ & $0.00\pm{0.00}$ & $-122.71\pm{68.24}$ & $1.17\pm{0.02}$ & $0.03\pm{0.01}$ & $2.37\pm{0.03}$ & $0.73\pm{0.23}$ \\
 & SACE & $\boldsymbol{1.00\pm{0.00}}$ & $\boldsymbol{1.00\pm{0.00}}$ & $0.64\pm{0.01}$ & $0.60\pm{0.55}$ & $\boldsymbol{80.45\pm{4.90}}$ & $\boldsymbol{1.01\pm{0.05}}$ & $\boldsymbol{0.06\pm{0.02}}$ & $3.65\pm{0.12}$ & $\boldsymbol{0.02\pm{0.00}}$ \\
 & DiCE & $\boldsymbol{1.00\pm{0.00}}$ & $0.63\pm{0.30}$ & $0.35\pm{0.03}$ & $0.00\pm{0.00}$ & $-1642.25\pm{1605.78}$ & $1.82\pm{0.03}$ & $-0.09\pm{0.01}$ & $2.81\pm{0.12}$ & $16.92\pm{1.62}$ \\
 & PPCEF & $\boldsymbol{1.00\pm{0.00}}$ & $\boldsymbol{1.00\pm{0.00}}$ & $\boldsymbol{1.00\pm{0.00}}$ & $\boldsymbol{1.00\pm{0.00}}$ & $64.52\pm{24.91}$ & $1.30\pm{0.03}$ & $0.00\pm{0.01}$ & $2.51\pm{0.37}$ & $13.66\pm{0.06}$ \\
 & Artelt & $0.80\pm{0.45}$ & $0.95\pm{0.05}$ & $\boldsymbol{1.00\pm{0.00}}$ & $0.00\pm{0.00}$ & $-4.83\times 10^{7}\pm{9.73\times 10^{7}}$ & $2.62\pm{0.14}$ & $0.03\pm{0.05}$ & $3.77\pm{0.13}$ & $28.83\pm{23.04}$ \\
 & CEGP & $\boldsymbol{1.00\pm{0.00}}$ & $0.95\pm{0.09}$ & $0.52\pm{0.05}$ & $0.00\pm{0.00}$ & $-342.05\pm{192.81}$ & $1.31\pm{0.04}$ & $-0.04\pm{0.01}$ & $1.57\pm{0.07}$ & $56.68\pm{2.12}$ \\
 & CEM & $\boldsymbol{1.00\pm{0.00}}$ & $0.95\pm{0.09}$ & $0.27\pm{0.05}$ & $0.00\pm{0.00}$ & $-1796.37\pm{1908.03}$ & $1.85\pm{0.17}$ & $-0.11\pm{0.02}$ & $2.64\pm{0.23}$ & $24.05\pm{0.78}$ \\
 & WACH & $\boldsymbol{1.00\pm{0.00}}$ & $\boldsymbol{1.00\pm{0.00}}$ & $\boldsymbol{1.00\pm{0.00}}$ & $0.00\pm{0.00}$ & $-966.89\pm{518.93}$ & $1.29\pm{0.04}$ & $-0.04\pm{0.01}$ & $1.37\pm{0.06}$ & $0.07\pm{0.00}$ \\
\midrule
\multirow{9}{*}{Moons} & CADEX & $\boldsymbol{1.00\pm{0.00}}$ & $\boldsymbol{1.00\pm{0.00}}$ & $\boldsymbol{1.00\pm{0.00}}$ & $0.03\pm{0.03}$ & $-0.98\pm{0.58}$ & $1.31\pm{0.05}$ & $0.00\pm{0.00}$ & $0.27\pm{0.01}$ & $1.53\pm{0.04}$ \\
 & CCHVAE & $\boldsymbol{1.00\pm{0.00}}$ & $\boldsymbol{1.00\pm{0.00}}$ & $\boldsymbol{1.00\pm{0.00}}$ & $0.00\pm{0.00}$ & $-1.38\pm{1.12}$ & $1.66\pm{0.06}$ & $0.02\pm{0.00}$ & $0.34\pm{0.02}$ & $0.31\pm{0.18}$ \\
 & SACE & $\boldsymbol{1.00\pm{0.00}}$ & $\boldsymbol{1.00\pm{0.00}}$ & $\boldsymbol{1.00\pm{0.00}}$ & $0.79\pm{0.45}$ & $\boldsymbol{1.32\pm{0.16}}$ & $1.18\pm{0.07}$ & $0.02\pm{0.01}$ & $0.44\pm{0.01}$ & $\boldsymbol{0.04\pm{0.00}}$ \\
 & DiCE & $\boldsymbol{1.00\pm{0.00}}$ & $0.98\pm{0.04}$ & $0.74\pm{0.02}$ & $0.19\pm{0.04}$ & $-6.44\pm{4.46}$ & $1.83\pm{0.05}$ & $-0.05\pm{0.01}$ & $0.59\pm{0.02}$ & $1.73\pm{0.05}$ \\
 & PPCEF & $\boldsymbol{1.00\pm{0.00}}$ & $\boldsymbol{1.00\pm{0.00}}$ & $\boldsymbol{1.00\pm{0.00}}$ & $\boldsymbol{0.90\pm{0.22}}$ & $1.30\pm{0.13}$ & $\boldsymbol{1.00\pm{0.01}}$ & $\boldsymbol{0.03\pm{0.00}}$ & $0.35\pm{0.04}$ & $39.03\pm{0.23}$ \\
 & Artelt & $\boldsymbol{1.00\pm{0.00}}$ & $\boldsymbol{1.00\pm{0.00}}$ & $\boldsymbol{1.00\pm{0.00}}$ & $0.04\pm{0.04}$ & $-1.13\pm{0.88}$ & $1.28\pm{0.02}$ & $-0.02\pm{0.01}$ & $0.28\pm{0.01}$ & $2.96\pm{0.23}$ \\
 & CEGP & $\boldsymbol{1.00\pm{0.00}}$ & $\boldsymbol{1.00\pm{0.00}}$ & $0.98\pm{0.01}$ & $0.05\pm{0.05}$ & $-0.75\pm{0.63}$ & $1.26\pm{0.05}$ & $0.00\pm{0.01}$ & $\boldsymbol{0.26\pm{0.01}}$ & $141.62\pm{3.13}$ \\
 & CEM & $\boldsymbol{1.00\pm{0.00}}$ & $\boldsymbol{1.00\pm{0.00}}$ & $0.59\pm{0.01}$ & $0.22\pm{0.04}$ & $-4.02\pm{1.68}$ & $1.67\pm{0.06}$ & $-0.05\pm{0.01}$ & $0.46\pm{0.01}$ & $60.42\pm{1.49}$ \\
 & WACH & $\boldsymbol{1.00\pm{0.00}}$ & $\boldsymbol{1.00\pm{0.00}}$ & $\boldsymbol{1.00\pm{0.00}}$ & $0.09\pm{0.06}$ & $-0.79\pm{0.58}$ & $1.36\pm{0.07}$ & $0.00\pm{0.00}$ & $0.30\pm{0.01}$ & $0.09\pm{0.01}$ \\
\midrule
\multirow{9}{*}{Wine} & CADEX & $\boldsymbol{1.00\pm{0.00}}$ & $\boldsymbol{1.00\pm{0.00}}$ & $\boldsymbol{1.00\pm{0.00}}$ & $0.02\pm{0.04}$ & $-3.39\pm{2.18}$ & $1.06\pm{0.04}$ & $0.05\pm{0.01}$ & $\boldsymbol{0.37\pm{0.02}}$ & $0.08\pm{0.01}$ \\
 & CCHVAE & $\boldsymbol{1.00\pm{0.00}}$ & $\boldsymbol{1.00\pm{0.00}}$ & $\boldsymbol{1.00\pm{0.00}}$ & $0.20\pm{0.45}$ & $2.96\pm{2.98}$ & $\boldsymbol{1.01\pm{0.01}}$ & $0.07\pm{0.01}$ & $0.86\pm{0.04}$ & $\boldsymbol{0.00\pm{0.00}}$ \\
 & SACE & $\boldsymbol{1.00\pm{0.00}}$ & $\boldsymbol{1.00\pm{0.00}}$ & $0.98\pm{0.02}$ & $0.00\pm{0.00}$ & $1.74\pm{2.11}$ & $1.27\pm{0.09}$ & $-0.03\pm{0.03}$ & $1.30\pm{0.11}$ & $\boldsymbol{0.00\pm{0.00}}$ \\
 & DiCE & $\boldsymbol{1.00\pm{0.00}}$ & $0.94\pm{0.09}$ & $0.23\pm{0.04}$ & $0.00\pm{0.00}$ & $-36.04\pm{28.43}$ & $1.37\pm{0.04}$ & $-0.02\pm{0.01}$ & $0.95\pm{0.05}$ & $0.43\pm{0.07}$ \\
 & PPCEF & $\boldsymbol{1.00\pm{0.00}}$ & $\boldsymbol{1.00\pm{0.00}}$ & $\boldsymbol{1.00\pm{0.00}}$ & $\boldsymbol{1.00\pm{0.00}}$ & $\boldsymbol{5.78\pm{1.13}}$ & $\boldsymbol{1.01\pm{0.02}}$ & $\boldsymbol{0.08\pm{0.01}}$ & $0.52\pm{0.07}$ & $8.21\pm{0.09}$ \\
 & Artelt & $0.80\pm{0.45}$ & $\boldsymbol{1.00\pm{0.00}}$ & $\boldsymbol{1.00\pm{0.00}}$ & $0.00\pm{0.00}$ & $-24.03\pm{19.19}$ & $1.25\pm{0.07}$ & $0.05\pm{0.05}$ & $0.73\pm{0.09}$ & $0.54\pm{0.06}$ \\
 & CEGP & $\boldsymbol{1.00\pm{0.00}}$ & $\boldsymbol{1.00\pm{0.00}}$ & $0.82\pm{0.05}$ & $0.06\pm{0.08}$ & $-2.34\pm{1.42}$ & $1.05\pm{0.04}$ & $0.05\pm{0.01}$ & $\boldsymbol{0.37\pm{0.05}}$ & $15.15\pm{2.42}$ \\
 & CEM & $\boldsymbol{1.00\pm{0.00}}$ & $\boldsymbol{1.00\pm{0.00}}$ & $0.34\pm{0.04}$ & $0.00\pm{0.00}$ & $-26.30\pm{9.67}$ & $1.33\pm{0.05}$ & $-0.03\pm{0.01}$ & $0.63\pm{0.10}$ & $6.56\pm{0.84}$ \\
 & WACH & $\boldsymbol{1.00\pm{0.00}}$ & $\boldsymbol{1.00\pm{0.00}}$ & $\boldsymbol{1.00\pm{0.00}}$ & $0.06\pm{0.08}$ & $-3.74\pm{3.13}$ & $1.07\pm{0.04}$ & $0.05\pm{0.01}$ & $0.42\pm{0.05}$ & $0.04\pm{0.00}$ \\
\midrule
\multirow{9}{*}{Audit} & CADEX & $\boldsymbol{1.00\pm{0.00}}$ & $\boldsymbol{1.00\pm{0.01}}$ & $\boldsymbol{1.00\pm{0.01}}$ & $0.08\pm{0.06}$ & $-360191.75\pm{757813.69}$ & $4.21\times 10^{7}\pm{9.31\times 10^{7}}$ & $0.05\pm{0.01}$ & $0.31\pm{0.02}$ & $0.30\pm{0.03}$ \\
 & CCHVAE & $\boldsymbol{1.00\pm{0.00}}$ & $\boldsymbol{1.00\pm{0.00}}$ & $\boldsymbol{1.00\pm{0.00}}$ & $0.45\pm{0.51}$ & $43.14\pm{6.25}$ & $100.11\pm{15.59}$ & $0.07\pm{0.01}$ & $1.36\pm{0.08}$ & $1.19\pm{0.08}$ \\
 & SACE & $\boldsymbol{1.00\pm{0.00}}$ & $\boldsymbol{1.00\pm{0.00}}$ & $0.61\pm{0.01}$ & $\boldsymbol{1.00\pm{0.00}}$ & $\boldsymbol{58.75\pm{3.11}}$ & $\boldsymbol{1.52\pm{1.09}}$ & $\boldsymbol{0.11\pm{0.01}}$ & $1.53\pm{0.13}$ & $\boldsymbol{0.03\pm{0.00}}$ \\
 & DiCE & $\boldsymbol{1.00\pm{0.00}}$ & $0.31\pm{0.03}$ & $0.02\pm{0.00}$ & $0.09\pm{0.09}$ & $-1154.88\pm{1945.34}$ & $3.75\times 10^{6}\pm{7.05\times 10^{6}}$ & $0.01\pm{0.01}$ & $0.77\pm{0.13}$ & $10.46\pm{0.57}$ \\
 & PPCEF & $\boldsymbol{1.00\pm{0.00}}$ & $\boldsymbol{1.00\pm{0.00}}$ & $\boldsymbol{1.00\pm{0.00}}$ & $0.99\pm{0.01}$ & $48.30\pm{7.23}$ & $1.20\times 10^{6}\pm{2.66\times 10^{6}}$ & $0.08\pm{0.01}$ & $0.99\pm{0.04}$ & $40.11\pm{0.15}$ \\
 & Artelt & $0.80\pm{0.45}$ & $0.98\pm{0.02}$ & $\boldsymbol{1.00\pm{0.00}}$ & $0.05\pm{0.05}$ & $-7232.86\pm{10576.85}$ & $2.52\pm{1.18}$ & $0.05\pm{0.07}$ & $0.52\pm{0.12}$ & $36.95\pm{19.88}$ \\
 & CEGP & $\boldsymbol{1.00\pm{0.00}}$ & $0.54\pm{0.06}$ & $0.23\pm{0.03}$ & $0.10\pm{0.10}$ & $-1092.96\pm{1981.30}$ & $7.00\times 10^{7}\pm{1.14\times 10^{8}}$ & $0.03\pm{0.01}$ & $0.46\pm{0.05}$ & $90.67\pm{5.54}$ \\
 & CEM & $\boldsymbol{1.00\pm{0.00}}$ & $0.02\pm{0.02}$ & $0.00\pm{0.00}$ & $0.13\pm{0.10}$ & $-1066.42\pm{1983.67}$ & $4.45\pm{4.04}$ & $0.01\pm{0.01}$ & $\boldsymbol{0.12\pm{0.20}}$ & $39.90\pm{2.83}$ \\
 & WACH & $\boldsymbol{1.00\pm{0.00}}$ & $\boldsymbol{1.00\pm{0.00}}$ & $\boldsymbol{1.00\pm{0.00}}$ & $0.06\pm{0.05}$ & $-2.32\times 10^{32}\pm{5.60\times 10^{32}}$ & $4.20\times 10^{7}\pm{9.30\times 10^{7}}$ & $0.05\pm{0.01}$ & $0.39\pm{0.11}$ & $0.18\pm{0.25}$ \\
\midrule
\multirow{9}{*}{HELOC} & CADEX & $\boldsymbol{1.00\pm{0.00}}$ & $\boldsymbol{1.00\pm{0.00}}$ & $\boldsymbol{1.00\pm{0.00}}$ & $0.34\pm{0.04}$ & $29.71\pm{0.51}$ & $1.13\pm{0.01}$ & $0.06\pm{0.01}$ & $0.12\pm{0.01}$ & $1.85\pm{0.08}$ \\
 & CCHVAE & $\boldsymbol{1.00\pm{0.00}}$ & $\boldsymbol{1.00\pm{0.00}}$ & $\boldsymbol{1.00\pm{0.00}}$ & $\boldsymbol{1.00\pm{0.00}}$ & $\boldsymbol{41.04\pm{1.51}}$ & $\boldsymbol{1.01\pm{0.01}}$ & $0.12\pm{0.00}$ & $0.50\pm{0.02}$ & $3.99\pm{2.07}$ \\
 & SACE & $\boldsymbol{1.00\pm{0.00}}$ & $\boldsymbol{1.00\pm{0.00}}$ & $0.82\pm{0.03}$ & $0.20\pm{0.45}$ & $35.19\pm{6.62}$ & $1.04\pm{0.04}$ & $0.07\pm{0.01}$ & $0.70\pm{0.03}$ & $3.27\pm{0.04}$ \\
 & DiCE & $\boldsymbol{1.00\pm{0.00}}$ & $0.93\pm{0.11}$ & $0.07\pm{0.00}$ & $0.01\pm{0.01}$ & $-149.95\pm{55.28}$ & $2.19\pm{0.22}$ & $0.04\pm{0.00}$ & $0.68\pm{0.03}$ & $45.23\pm{0.20}$ \\
 & PPCEF & $\boldsymbol{1.00\pm{0.00}}$ & $\boldsymbol{1.00\pm{0.00}}$ & $\boldsymbol{1.00\pm{0.00}}$ & $\boldsymbol{1.00\pm{0.00}}$ & $38.38\pm{0.99}$ & $1.08\pm{0.01}$ & $0.07\pm{0.01}$ & $0.23\pm{0.02}$ & $136.59\pm{1.53}$ \\
 & Artelt & $0.00\pm{0.00}$ & $\boldsymbol{1.00\pm{0.00}}$ & $\boldsymbol{1.00\pm{0.00}}$ & $0.00\pm{0.00}$ & -- & -- & $\boldsymbol{0.15\pm{0.00}}$ & -- & $161.97\pm{25.25}$ \\
 & CEGP & $\boldsymbol{1.00\pm{0.00}}$ & $\boldsymbol{1.00\pm{0.00}}$ & $0.43\pm{0.01}$ & $0.37\pm{0.04}$ & $30.23\pm{2.13}$ & $1.14\pm{0.01}$ & $0.05\pm{0.00}$ & $\boldsymbol{0.11\pm{0.01}}$ & $1497.32\pm{13.02}$ \\
 & CEM & $\boldsymbol{1.00\pm{0.00}}$ & $\boldsymbol{1.00\pm{0.00}}$ & $0.23\pm{0.02}$ & $0.16\pm{0.03}$ & $19.31\pm{5.49}$ & $1.28\pm{0.03}$ & $0.02\pm{0.01}$ & $0.19\pm{0.01}$ & $610.19\pm{9.90}$ \\
 & WACH & $\boldsymbol{1.00\pm{0.00}}$ & $\boldsymbol{1.00\pm{0.00}}$ & $\boldsymbol{1.00\pm{0.00}}$ & $0.20\pm{0.05}$ & $25.05\pm{2.48}$ & $1.18\pm{0.03}$ & $0.04\pm{0.01}$ & $0.18\pm{0.03}$ & $\boldsymbol{0.03\pm{0.00}}$ \\
\bottomrule
\end{tabular}
\end{scriptsize}
\end{sc}
\end{center}
\end{table*}

\subsection{Regression Methods}

This subsection reports results for counterfactual explanation methods applied to regression tasks. Validity for regression is measured as the mean absolute error (MAE) between the predicted value of the counterfactual and the desired target value, rather than as a binary classification success rate. The evaluated methods include CEARM and Wachter (WACH), tested on five datasets: Synthetic, Concrete, Diabetes, Yacht, and SCM20D. Results are reported for both Linear Regression and MLP Regressor models. Tables~\ref{tab:regression_all_lr} and~\ref{tab:regression_all_dnn} report the performance metrics for Linear Regression and MLP backbones, respectively.

CEARM is not applicable to the SCM20D dataset (multi-output regression with 16 targets), leaving Wachter as the only method evaluated in this setting. On the remaining four datasets, both methods achieve comparable MAE values, indicating similar effectiveness in reaching the target prediction. For instance, on Synthetic both methods achieve MAE of approximately 0.05 under both backbones.

However, Wachter consistently achieves substantially lower proximity values. Its L2 distances are typically several times smaller than CEARM's across all datasets and backbones (e.g., 0.14 vs.\ 0.37 on Synthetic with MLP; 0.09 vs.\ 1.16 on Concrete with MLP). This indicates that Wachter produces counterfactuals with more efficient perturbations while achieving comparable target accuracy.

In terms of plausibility, Wachter achieves better outlier scores across nearly all configurations. Its LOF values are consistently close to 1.0 (e.g., 1.04--1.17), while CEARM produces higher LOF values (1.14--2.27), suggesting that CEARM's counterfactuals are more likely to lie in low-density regions. Wachter also achieves higher Isolation Forest scores and substantially better log-density values. On Concrete with MLP, Wachter achieves a log-density of 5.70 compared to $-152.55$ for CEARM. Under the LR backbone, CEARM achieves higher probabilistic plausibility only on Synthetic (0.16 vs.\ 0.03 for Wachter), while under MLP, Wachter achieves higher probabilistic plausibility on Concrete (0.24 vs.\ 0.00), Diabetes (0.10 vs.\ 0.00), and Yacht (0.10 vs.\ 0.01), indicating that its gradient-based optimization is more effective at finding plausible counterfactuals for nonlinear models.

Wachter is also significantly faster, completing in 3--14 seconds compared to CEARM's 24--100 seconds, due to its simpler gradient-based optimization.

\begin{table*}[ht!]
\centering
\caption{Comparative regression results for the \textbf{Linear Regression} model. All performance metrics.}
\label{tab:regression_all_lr}
\begin{center}
\begin{sc}
\begin{tiny}
\begin{tabular}{@{}llrrrrrrrrr@{}}
\toprule
Dataset & Method & MAE $\downarrow$ & PP $\uparrow$ & L2 $\downarrow$ & LOF $\downarrow$ & IF $\uparrow$ & LD $\uparrow$ & L1 $\downarrow$ & Time $\downarrow$ \\
\midrule
\multirow[t]{3}{*}{Synthetic} & CEARM & 0.05$\pm$0.01 & 0.16$\pm$0.03 & 0.38$\pm$0.03 & 2.06$\pm$0.09 & -0.09$\pm$0.01 & -83.59$\pm$52.05 & 0.48$\pm$0.04 & 66.84$\pm$2.10 \\
 & WACH & 0.06$\pm$0.01 & 0.03$\pm$0.05 & \textbf{0.15$\pm$0.02} & \textbf{1.08$\pm$0.02} & \textbf{0.02$\pm$0.01} & 0.87$\pm$2.27 & \textbf{0.21$\pm$0.03} & \textbf{3.13$\pm$0.28} \\
\midrule
\multirow[t]{3}{*}{Concrete} & CEARM & 0.03$\pm$0.00 & 0.00$\pm$0.00 & 1.13$\pm$0.02 & 2.21$\pm$0.04 & -0.11$\pm$0.00 & -2880.93$\pm$3261.61 & 2.63$\pm$0.05 & 100.42$\pm$5.76 \\
 & WACH & 0.05$\pm$0.00 & 0.00$\pm$0.00 & \textbf{0.16$\pm$0.00} & \textbf{1.17$\pm$0.02} & \textbf{0.00$\pm$0.01} & 0.45$\pm$2.17 & \textbf{0.36$\pm$0.01} & \textbf{3.15$\pm$0.27} \\
\midrule
\multirow[t]{3}{*}{Diabetes} & CEARM & 0.04$\pm$0.01 & 0.00$\pm$0.00 & 1.25$\pm$0.03 & 1.93$\pm$0.04 & -0.10$\pm$0.01 & -229.94$\pm$170.66 & 3.29$\pm$0.11 & 44.70$\pm$1.29 \\
 & WACH & 0.07$\pm$0.01 & 0.00$\pm$0.00 & \textbf{0.18$\pm$0.01} & 1.09$\pm$0.01 & 0.03$\pm$0.01 & -3.11$\pm$4.60 & \textbf{0.44$\pm$0.06} & \textbf{3.57$\pm$0.09} \\
\midrule
\multirow[t]{3}{*}{Yacht} & CEARM & 0.08$\pm$0.02 & 0.00$\pm$0.01 & 1.02$\pm$0.04 & 1.17$\pm$0.02 & -0.06$\pm$0.01 & -8988.85$\pm$12145.27 & 2.15$\pm$0.11 & 25.23$\pm$1.06 \\
 & WACH & 0.08$\pm$0.01 & 0.00$\pm$0.01 & \textbf{0.22$\pm$0.01} & 1.05$\pm$0.02 & \textbf{-0.03$\pm$0.01} & -14.39$\pm$11.12 & \textbf{0.34$\pm$0.05} & \textbf{3.48$\pm$0.07} \\
\midrule
\multirow[t]{2}{*}{SCM20D} & WACH & 0.07$\pm$0.00 & 0.00$\pm$0.00 & \textbf{0.18$\pm$0.00} & \textbf{1.04$\pm$0.00} & \textbf{0.01$\pm$0.00} & 26.17$\pm$7.72 & \textbf{1.13$\pm$0.03} & \textbf{5.87$\pm$0.49} \\
\bottomrule
\end{tabular}
\end{tiny}
\end{sc}
\end{center}
\end{table*}

\begin{table*}[ht!]
\centering
\caption{Comparative regression results for the \textbf{Multilayer Perceptron} model. All performance metrics.}
\label{tab:regression_all_dnn}
\begin{center}
\begin{sc}
\begin{tiny}
\begin{tabular}{@{}llrrrrrrrrr@{}}
\toprule
Dataset & Method & MAE $\downarrow$ & PP $\uparrow$ & L2 $\downarrow$ & LOF $\downarrow$ & IF $\uparrow$ & LD $\uparrow$ & L1 $\downarrow$ & Time $\downarrow$ \\
\midrule
\multirow[t]{3}{*}{Synthetic} & CEARM & 0.05$\pm$0.01 & 0.20$\pm$0.02 & 0.37$\pm$0.02 & 2.11$\pm$0.10 & -0.09$\pm$0.01 & -49.49$\pm$47.20 & 0.47$\pm$0.02 & 66.91$\pm$0.86 \\
 & WACH & 0.05$\pm$0.00 & 0.01$\pm$0.02 & \textbf{0.14$\pm$0.01} & 1.09$\pm$0.02 & \textbf{0.02$\pm$0.00} & 1.35$\pm$1.07 & \textbf{0.20$\pm$0.02} & \textbf{12.28$\pm$1.43} \\
\midrule
\multirow[t]{3}{*}{Concrete} & CEARM & 0.04$\pm$0.01 & 0.00$\pm$0.00 & 1.16$\pm$0.02 & 2.27$\pm$0.04 & -0.12$\pm$0.00 & -152.55$\pm$322.80 & 2.72$\pm$0.05 & 94.28$\pm$1.11 \\
 & WACH & 0.03$\pm$0.00 & 0.24$\pm$0.02 & \textbf{0.09$\pm$0.00} & 1.13$\pm$0.02 & 0.02$\pm$0.01 & 5.70$\pm$0.44 & \textbf{0.21$\pm$0.01} & \textbf{12.17$\pm$1.55} \\
\midrule
\multirow[t]{3}{*}{Diabetes} & CEARM & 0.05$\pm$0.02 & 0.00$\pm$0.00 & 1.21$\pm$0.03 & 1.89$\pm$0.04 & -0.09$\pm$0.00 & -467.25$\pm$493.84 & 3.18$\pm$0.11 & 43.92$\pm$1.87 \\
 & WACH & 0.05$\pm$0.01 & 0.10$\pm$0.07 & \textbf{0.15$\pm$0.01} & 1.09$\pm$0.01 & 0.03$\pm$0.00 & 3.67$\pm$2.46 & \textbf{0.40$\pm$0.02} & \textbf{4.92$\pm$0.26} \\
\midrule
\multirow[t]{3}{*}{Yacht} & CEARM & 0.13$\pm$0.05 & 0.01$\pm$0.01 & 1.04$\pm$0.02 & 1.14$\pm$0.04 & -0.05$\pm$0.01 & -837.38$\pm$1115.65 & 2.18$\pm$0.07 & 24.31$\pm$1.67 \\
 & WACH & 0.12$\pm$0.02 & 0.10$\pm$0.04 & \textbf{0.05$\pm$0.01} & 1.05$\pm$0.01 & -0.03$\pm$0.01 & -70.40$\pm$32.08 & \textbf{0.07$\pm$0.01} & \textbf{4.81$\pm$0.19} \\
\midrule
\multirow[t]{2}{*}{SCM20D} & WACH & 0.15$\pm$0.01 & 0.00$\pm$0.00 & \textbf{0.11$\pm$0.01} & \textbf{1.04$\pm$0.00} & \textbf{0.01$\pm$0.00} & 32.87$\pm$10.08 & \textbf{0.67$\pm$0.09} & \textbf{14.40$\pm$2.43} \\
\bottomrule
\end{tabular}
\end{tiny}
\end{sc}
\end{center}
\end{table*}

\subsection{Global Methods}

This subsection reports results for global counterfactual explanation methods, which learn a single transformation or shift vector applied uniformly across all instances. The evaluated methods include AReS, GLOBE-CE, and GLANCE configured with a single group (GLANCE). Results are reported across all 13 classification datasets and for both MLP and Logistic Regression backbones. Tables~\ref{tab:global_metrics_mlp} and~\ref{tab:global_metrics_lr} report all metrics for MLP and Logistic Regression, respectively.

A notable observation is that AReS fails to produce results (indicated by --) on many datasets, including Audit, Blobs, Credit Default, Digits, Law, Moons, and Wine under both backbones. Additionally, AReS fails on Lending Club under Logistic Regression. This substantially limits its applicability compared to the other global methods. Similarly, GLOBE-CE fails on the Digits dataset under both backbones, likely due to the high dimensionality (64 features) making it difficult to find a single effective global shift.

Where all three methods produce results, clear trade-offs emerge. GLOBE-CE generally achieves the lowest proximity values, often by a large margin. For example, on Adult Census it achieves near-zero distances across all proximity metrics under both backbones, and on Give Me Some Credit it achieves L2-Hamming distances of 0.01 compared to 0.86 for GLANCE (MLP). GLOBE-CE also tends to achieve high or perfect validity on most datasets where it succeeds (e.g., 1.00 on Adult Census, Give Me Some Credit, Moons, Law). However, on some datasets it achieves lower validity (e.g., 0.67 on Bank Marketing, 0.75 on Credit Default with MLP), indicating that a single global shift may not always suffice.

GLANCE generally achieves the highest validity among the three methods, reaching perfect or near-perfect validity on many datasets (e.g., 1.00 on Adult Census, Bank Marketing, and Give Me Some Credit with both backbones). However, it shows reduced validity on some datasets such as Audit (0.40 with MLP), Moons (0.69 with MLP), and Wine (0.80 with MLP), and this comes at the cost of substantially larger perturbations. AReS, when it succeeds, generally achieves low proximity but often with very low validity (e.g., 0.17 on Adult Census, 0.06 on Bank Marketing with MLP), suggesting that its rule-based recourse often fails to flip predictions.

In terms of plausibility, no single method dominates. AReS achieves the highest probabilistic plausibility on several datasets where it produces results (e.g., 0.50 on Give Me Some Credit, 0.41 on German Credit). GLOBE-CE is consistently the fastest method, typically completing in under 10 seconds, while GLANCE computation time varies widely from under 3 seconds on smaller datasets (e.g., Digits, German Credit) to over 250 seconds on larger ones (e.g., Adult Census, Credit Default).

\begin{table*}[ht]
\centering
\caption{Global methods: metrics (MLP).}
\label{tab:global_metrics_mlp}
\begin{center}
\begin{sc}
\begin{scriptsize}
\begin{tabular}{l|l|rrrrrrrrr}
\toprule
 & Method & Cov.$\uparrow$ & Valid.$\uparrow$ & Sparse.$\uparrow$ & Prob. Plaus.$\uparrow$ & Log Dens.$\uparrow$ & LOF$\downarrow$ & IsoForest$\uparrow$ & L2-Ham.$\downarrow$ & Time(s)$\downarrow$ \\
\midrule
\multirow{3}{*}{Adult Census} & GLANCE & $\boldsymbol{1.00\pm{0.00}}$ & $\boldsymbol{1.00\pm{0.00}}$ & $\boldsymbol{0.85\pm{0.03}}$ & $0.35\pm{0.07}$ & $-247.54\pm{75.60}$ & $4.00\pm{0.16}$ & $0.10\pm{0.01}$ & $0.84\pm{0.03}$ & $259.98\pm{1.95}$ \\
 & AReS & $\boldsymbol{1.00\pm{0.00}}$ & $0.17\pm{0.11}$ & $0.01\pm{0.00}$ & $0.34\pm{0.20}$ & $-399.45\pm{353.04}$ & $6.56\pm{2.49}$ & $0.11\pm{0.00}$ & $0.01\pm{0.01}$ & $97.73\pm{21.67}$ \\
 & GLOBE-CE & $\boldsymbol{1.00\pm{0.00}}$ & $\boldsymbol{1.00\pm{0.00}}$ & $0.01\pm{0.00}$ & $\boldsymbol{0.49\pm{0.15}}$ & $\boldsymbol{-241.22\pm{14.60}}$ & $\boldsymbol{3.16\pm{0.06}}$ & $\boldsymbol{0.12\pm{0.00}}$ & $\boldsymbol{0.00\pm{0.00}}$ & $\boldsymbol{7.39\pm{0.23}}$ \\
\midrule
\multirow{3}{*}{Audit} & GLANCE & $\boldsymbol{1.00\pm{0.00}}$ & $0.40\pm{0.03}$ & $\boldsymbol{1.00\pm{0.00}}$ & $0.00\pm{0.00}$ & $\boldsymbol{18.72\pm{9.43}}$ & $\boldsymbol{114.53\pm{26.35}}$ & $-0.09\pm{0.04}$ & $\boldsymbol{0.35\pm{0.04}}$ & $28.28\pm{0.05}$ \\
 & AReS & -- & -- & -- & -- & -- & -- & -- & -- & -- \\
 & GLOBE-CE & $\boldsymbol{1.00\pm{0.00}}$ & $\boldsymbol{1.00\pm{0.00}}$ & $0.44\pm{0.02}$ & $\boldsymbol{0.03\pm{0.02}}$ & $-291.16\pm{473.23}$ & $119.56\pm{125.84}$ & $\boldsymbol{0.11\pm{0.04}}$ & $0.91\pm{0.22}$ & $\boldsymbol{0.38\pm{0.01}}$ \\
\midrule
\multirow{3}{*}{Bank Marketing} & GLANCE & $\boldsymbol{1.00\pm{0.00}}$ & $\boldsymbol{0.99\pm{0.02}}$ & $\boldsymbol{1.00\pm{0.01}}$ & $0.00\pm{0.01}$ & $-72.77\pm{4.33}$ & $\boldsymbol{3.37\pm{0.23}}$ & $0.01\pm{0.03}$ & $0.96\pm{0.01}$ & $96.32\pm{8.41}$ \\
 & AReS & $\boldsymbol{1.00\pm{0.00}}$ & $0.06\pm{0.09}$ & $0.05\pm{0.02}$ & $0.07\pm{0.09}$ & $\boldsymbol{-55.23\pm{5.51}}$ & $3.41\pm{1.46}$ & $0.03\pm{0.02}$ & $\boldsymbol{0.05\pm{0.03}}$ & $201.45\pm{33.89}$ \\
 & GLOBE-CE & $\boldsymbol{1.00\pm{0.00}}$ & $0.67\pm{0.46}$ & $0.08\pm{0.04}$ & $\boldsymbol{0.37\pm{0.30}}$ & -- & $6.35\pm{6.64}$ & $\boldsymbol{0.04\pm{0.02}}$ & $0.08\pm{0.07}$ & $\boldsymbol{5.98\pm{0.13}}$ \\
\midrule
\multirow{3}{*}{Blobs} & GLANCE & $\boldsymbol{1.00\pm{0.00}}$ & $0.98\pm{0.02}$ & $\boldsymbol{1.00\pm{0.00}}$ & $\boldsymbol{0.35\pm{0.11}}$ & $\boldsymbol{0.66\pm{0.28}}$ & $\boldsymbol{1.26\pm{0.06}}$ & $\boldsymbol{-0.01\pm{0.01}}$ & $0.47\pm{0.05}$ & $100.57\pm{13.67}$ \\
 & AReS & -- & -- & -- & -- & -- & -- & -- & -- & -- \\
 & GLOBE-CE & $\boldsymbol{1.00\pm{0.00}}$ & $\boldsymbol{1.00\pm{0.00}}$ & $\boldsymbol{1.00\pm{0.00}}$ & $0.00\pm{0.00}$ & $-7.16\pm{2.42}$ & $2.14\pm{0.31}$ & $-0.08\pm{0.02}$ & $\boldsymbol{0.35\pm{0.05}}$ & $\boldsymbol{0.39\pm{0.01}}$ \\
\midrule
\multirow{3}{*}{Credit Default} & GLANCE & $\boldsymbol{1.00\pm{0.00}}$ & $\boldsymbol{0.97\pm{0.02}}$ & $\boldsymbol{1.00\pm{0.00}}$ & $\boldsymbol{0.00\pm{0.00}}$ & -- & $\boldsymbol{8.33\pm{0.10}}$ & $-0.05\pm{0.01}$ & $0.87\pm{0.01}$ & $259.52\pm{45.28}$ \\
 & AReS & -- & -- & -- & -- & -- & -- & -- & -- & -- \\
 & GLOBE-CE & $\boldsymbol{1.00\pm{0.00}}$ & $0.75\pm{0.50}$ & $0.04\pm{0.02}$ & $\boldsymbol{0.00\pm{0.00}}$ & $\boldsymbol{-5685.85\pm{6424.64}}$ & $207.52\pm{399.88}$ & $\boldsymbol{0.07\pm{0.01}}$ & $\boldsymbol{0.10\pm{0.05}}$ & $\boldsymbol{5.89\pm{0.09}}$ \\
\midrule
\multirow{3}{*}{Digits} & GLANCE & $\boldsymbol{1.00\pm{0.00}}$ & $\boldsymbol{0.74\pm{0.04}}$ & $\boldsymbol{0.91\pm{0.06}}$ & $\boldsymbol{0.00\pm{0.00}}$ & $\boldsymbol{-5585.73\pm{7131.33}}$ & $\boldsymbol{1.55\pm{0.06}}$ & $\boldsymbol{-0.07\pm{0.02}}$ & $\boldsymbol{1.77\pm{0.13}}$ & $\boldsymbol{2.54\pm{0.21}}$ \\
 & AReS & -- & -- & -- & -- & -- & -- & -- & -- & -- \\
 & GLOBE-CE & -- & $0.00\pm{0.00}$ & -- & -- & -- & -- & -- & -- & -- \\
\midrule
\multirow{3}{*}{German Credit} & GLANCE & $\boldsymbol{1.00\pm{0.00}}$ & $\boldsymbol{0.80\pm{0.07}}$ & $\boldsymbol{0.70\pm{0.09}}$ & $0.17\pm{0.14}$ & -- & $1.07\pm{0.01}$ & $-0.03\pm{0.01}$ & $0.64\pm{0.08}$ & $2.01\pm{0.92}$ \\
 & AReS & $\boldsymbol{1.00\pm{0.00}}$ & $0.29\pm{0.12}$ & $0.02\pm{0.01}$ & $\boldsymbol{0.41\pm{0.15}}$ & $\boldsymbol{-60.31\pm{2.58}}$ & $\boldsymbol{1.05\pm{0.02}}$ & $\boldsymbol{0.02\pm{0.02}}$ & $\boldsymbol{0.04\pm{0.00}}$ & $2.15\pm{1.49}$ \\
 & GLOBE-CE & $\boldsymbol{1.00\pm{0.00}}$ & $0.49\pm{0.13}$ & $0.08\pm{0.02}$ & $0.22\pm{0.19}$ & -- & $7.43\pm{10.23}$ & $\boldsymbol{0.02\pm{0.01}}$ & $0.09\pm{0.05}$ & $\boldsymbol{0.44\pm{0.00}}$ \\
\midrule
\multirow{3}{*}{GMC} & GLANCE & $\boldsymbol{1.00\pm{0.00}}$ & $\boldsymbol{1.00\pm{0.00}}$ & $\boldsymbol{1.00\pm{0.00}}$ & $0.00\pm{0.00}$ & $-27.54\pm{10.15}$ & $5.27\pm{0.32}$ & $-0.02\pm{0.01}$ & $0.86\pm{0.00}$ & $62.27\pm{1.92}$ \\
 & AReS & $\boldsymbol{1.00\pm{0.00}}$ & $0.88\pm{0.10}$ & $0.05\pm{0.01}$ & $\boldsymbol{0.50\pm{0.15}}$ & $\boldsymbol{-14.81\pm{5.71}}$ & $\boldsymbol{1.20\pm{0.04}}$ & $0.08\pm{0.02}$ & $0.05\pm{0.00}$ & $15.69\pm{9.07}$ \\
 & GLOBE-CE & $\boldsymbol{1.00\pm{0.00}}$ & $\boldsymbol{1.00\pm{0.00}}$ & $0.05\pm{0.01}$ & $0.42\pm{0.35}$ & $-15.17\pm{6.03}$ & $1.49\pm{0.67}$ & $\boldsymbol{0.11\pm{0.01}}$ & $\boldsymbol{0.01\pm{0.01}}$ & $\boldsymbol{1.92\pm{0.07}}$ \\
\midrule
\multirow{3}{*}{HELOC} & GLANCE & $\boldsymbol{1.00\pm{0.00}}$ & $0.97\pm{0.01}$ & $\boldsymbol{1.00\pm{0.00}}$ & $0.00\pm{0.00}$ & $\boldsymbol{-26269.57\pm{59527.61}}$ & $6.91\times 10^{8}\pm{8.77\times 10^{7}}$ & $0.00\pm{0.01}$ & $0.69\pm{0.05}$ & $63.73\pm{12.21}$ \\
 & AReS & $\boldsymbol{1.00\pm{0.00}}$ & $0.13\pm{0.06}$ & $0.24\pm{0.01}$ & $0.04\pm{0.03}$ & $-522982.51\pm{923938.83}$ & $\boldsymbol{38.22\pm{4.51}}$ & $\boldsymbol{0.04\pm{0.00}}$ & $\boldsymbol{0.18\pm{0.09}}$ & $\boldsymbol{0.20\pm{0.13}}$ \\
 & GLOBE-CE & $\boldsymbol{1.00\pm{0.00}}$ & $\boldsymbol{1.00\pm{0.00}}$ & $0.27\pm{0.01}$ & $\boldsymbol{0.13\pm{0.03}}$ & $-1.10\times 10^{14}\pm{2.38\times 10^{14}}$ & $5.25\times 10^{8}\pm{4.96\times 10^{8}}$ & $0.03\pm{0.00}$ & $0.36\pm{0.07}$ & $1.28\pm{0.04}$ \\
\midrule
\multirow{3}{*}{Law} & GLANCE & $\boldsymbol{1.00\pm{0.00}}$ & $0.93\pm{0.01}$ & $\boldsymbol{1.00\pm{0.00}}$ & $\boldsymbol{0.40\pm{0.05}}$ & $\boldsymbol{-15.39\pm{0.22}}$ & $1.70\pm{0.08}$ & $-0.01\pm{0.00}$ & $0.88\pm{0.01}$ & $27.75\pm{4.53}$ \\
 & AReS & -- & -- & -- & -- & -- & -- & -- & -- & -- \\
 & GLOBE-CE & $\boldsymbol{1.00\pm{0.00}}$ & $\boldsymbol{1.00\pm{0.00}}$ & $0.19\pm{0.01}$ & $0.21\pm{0.16}$ & $-18.21\pm{2.38}$ & $\boldsymbol{1.51\pm{0.05}}$ & $\boldsymbol{0.02\pm{0.00}}$ & $\boldsymbol{0.06\pm{0.00}}$ & $\boldsymbol{0.80\pm{0.01}}$ \\
\midrule
\multirow{3}{*}{Lending Club} & GLANCE & $\boldsymbol{1.00\pm{0.00}}$ & $\boldsymbol{0.99\pm{0.01}}$ & $\boldsymbol{1.00\pm{0.00}}$ & $0.00\pm{0.00}$ & -- & $2.84\pm{0.17}$ & $-0.04\pm{0.01}$ & $0.83\pm{0.02}$ & $87.66\pm{12.53}$ \\
 & AReS & $\boldsymbol{1.00\pm{0.00}}$ & $0.30\pm{0.47}$ & $0.08\pm{0.02}$ & $0.22\pm{0.40}$ & $-49.30\pm{13.73}$ & $\boldsymbol{1.84\pm{1.04}}$ & $0.02\pm{0.03}$ & $\boldsymbol{0.08\pm{0.06}}$ & $\boldsymbol{0.06\pm{0.04}}$ \\
 & GLOBE-CE & $\boldsymbol{1.00\pm{0.00}}$ & $0.75\pm{0.05}$ & $0.18\pm{0.01}$ & $\boldsymbol{0.34\pm{0.25}}$ & $\boldsymbol{-36.09\pm{6.59}}$ & $5.24\pm{0.45}$ & $\boldsymbol{0.04\pm{0.00}}$ & $\boldsymbol{0.08\pm{0.02}}$ & $1.84\pm{0.04}$ \\
\midrule
\multirow{3}{*}{Moons} & GLANCE & $\boldsymbol{1.00\pm{0.00}}$ & $0.69\pm{0.03}$ & $\boldsymbol{1.00\pm{0.00}}$ & $\boldsymbol{0.19\pm{0.05}}$ & $-14.29\pm{9.22}$ & $\boldsymbol{1.47\pm{0.05}}$ & $\boldsymbol{-0.01\pm{0.00}}$ & $0.41\pm{0.03}$ & $41.72\pm{3.94}$ \\
 & AReS & -- & -- & -- & -- & -- & -- & -- & -- & -- \\
 & GLOBE-CE & $\boldsymbol{1.00\pm{0.00}}$ & $\boldsymbol{1.00\pm{0.00}}$ & $\boldsymbol{1.00\pm{0.00}}$ & $0.00\pm{0.00}$ & $\boldsymbol{-12.46\pm{7.25}}$ & $2.20\pm{0.18}$ & $-0.07\pm{0.01}$ & $\boldsymbol{0.36\pm{0.04}}$ & $\boldsymbol{0.38\pm{0.00}}$ \\
\midrule
\multirow{3}{*}{Wine} & GLANCE & $\boldsymbol{1.00\pm{0.00}}$ & $0.80\pm{0.08}$ & $\boldsymbol{0.72\pm{0.07}}$ & $\boldsymbol{0.00\pm{0.00}}$ & $\boldsymbol{-20.75\pm{8.25}}$ & $\boldsymbol{1.25\pm{0.03}}$ & $\boldsymbol{0.00\pm{0.01}}$ & $\boldsymbol{0.91\pm{0.04}}$ & $1.51\pm{1.57}$ \\
 & AReS & -- & -- & -- & -- & -- & -- & -- & -- & -- \\
 & GLOBE-CE & $\boldsymbol{1.00\pm{0.00}}$ & $\boldsymbol{1.00\pm{0.00}}$ & $0.57\pm{0.01}$ & $\boldsymbol{0.00\pm{0.00}}$ & $-94.86\pm{48.69}$ & $2.03\pm{0.27}$ & $-0.01\pm{0.01}$ & $1.48\pm{0.23}$ & $\boldsymbol{0.33\pm{0.00}}$ \\
\bottomrule
\end{tabular}
\end{scriptsize}
\end{sc}
\end{center}
\end{table*}

\begin{table*}[ht]
\centering
\caption{Global methods: metrics (LR).}
\label{tab:global_metrics_lr}
\begin{center}
\begin{sc}
\begin{scriptsize}
\begin{tabular}{l|l|rrrrrrrrr}
\toprule
 & Method & Cov.$\uparrow$ & Valid.$\uparrow$ & Sparse.$\uparrow$ & Prob. Plaus.$\uparrow$ & Log Dens.$\uparrow$ & LOF$\downarrow$ & IsoForest$\uparrow$ & L2-Ham.$\downarrow$ & Time(s)$\downarrow$ \\
\midrule
\multirow{3}{*}{Adult Census} & GLANCE & $\boldsymbol{1.00\pm{0.00}}$ & $\boldsymbol{1.00\pm{0.00}}$ & $\boldsymbol{0.84\pm{0.03}}$ & $0.37\pm{0.06}$ & $\boldsymbol{-231.69\pm{40.35}}$ & $4.10\pm{0.15}$ & $0.10\pm{0.01}$ & $0.83\pm{0.03}$ & $257.83\pm{0.50}$ \\
 & AReS & $\boldsymbol{1.00\pm{0.00}}$ & $0.24\pm{0.14}$ & $0.01\pm{0.00}$ & $0.33\pm{0.26}$ & $-268.48\pm{27.85}$ & $4.81\pm{3.47}$ & $\boldsymbol{0.12\pm{0.01}}$ & $0.01\pm{0.01}$ & $80.82\pm{10.46}$ \\
 & GLOBE-CE & $\boldsymbol{1.00\pm{0.00}}$ & $\boldsymbol{1.00\pm{0.00}}$ & $0.01\pm{0.00}$ & $\boldsymbol{0.38\pm{0.19}}$ & $-247.90\pm{12.00}$ & $\boldsymbol{3.24\pm{0.05}}$ & $\boldsymbol{0.12\pm{0.00}}$ & $\boldsymbol{0.00\pm{0.00}}$ & $\boldsymbol{5.55\pm{0.10}}$ \\
\midrule
\multirow{3}{*}{Audit} & GLANCE & $\boldsymbol{1.00\pm{0.00}}$ & $\boldsymbol{1.00\pm{0.00}}$ & $\boldsymbol{1.00\pm{0.00}}$ & $0.00\pm{0.00}$ & $\boldsymbol{21.14\pm{5.87}}$ & $\boldsymbol{94.75\pm{26.62}}$ & $-0.10\pm{0.04}$ & $\boldsymbol{0.26\pm{0.00}}$ & $28.00\pm{0.07}$ \\
 & AReS & -- & -- & -- & -- & -- & -- & -- & -- & -- \\
 & GLOBE-CE & $\boldsymbol{1.00\pm{0.00}}$ & $\boldsymbol{1.00\pm{0.00}}$ & $0.46\pm{0.03}$ & $\boldsymbol{0.05\pm{0.05}}$ & $-17.72\pm{33.55}$ & $102.60\pm{36.96}$ & $\boldsymbol{0.10\pm{0.03}}$ & $0.27\pm{0.06}$ & $\boldsymbol{0.13\pm{0.00}}$ \\
\midrule
\multirow{3}{*}{Bank Marketing} & GLANCE & $\boldsymbol{1.00\pm{0.00}}$ & $\boldsymbol{1.00\pm{0.00}}$ & $\boldsymbol{1.00\pm{0.00}}$ & $\boldsymbol{0.18\pm{0.14}}$ & $\boldsymbol{-50.85\pm{9.93}}$ & $\boldsymbol{2.77\pm{0.08}}$ & $0.03\pm{0.01}$ & $0.94\pm{0.00}$ & $89.95\pm{0.19}$ \\
 & AReS & $\boldsymbol{1.00\pm{0.00}}$ & $0.09\pm{0.12}$ & $0.04\pm{0.02}$ & $0.07\pm{0.05}$ & $-56.45\pm{12.24}$ & $3.53\pm{1.61}$ & $\boldsymbol{0.04\pm{0.02}}$ & $\boldsymbol{0.03\pm{0.02}}$ & $194.48\pm{22.55}$ \\
 & GLOBE-CE & $\boldsymbol{1.00\pm{0.00}}$ & $0.85\pm{0.11}$ & $0.10\pm{0.03}$ & $0.06\pm{0.08}$ & $-5.94\times 10^{12}\pm{1.33\times 10^{13}}$ & $5.22\pm{1.92}$ & $0.02\pm{0.01}$ & $0.11\pm{0.06}$ & $\boldsymbol{3.77\pm{0.12}}$ \\
\midrule
\multirow{3}{*}{Blobs} & GLANCE & $\boldsymbol{1.00\pm{0.00}}$ & $0.99\pm{0.01}$ & $\boldsymbol{1.00\pm{0.00}}$ & $\boldsymbol{0.45\pm{0.09}}$ & $\boldsymbol{0.98\pm{0.39}}$ & $\boldsymbol{1.20\pm{0.07}}$ & $\boldsymbol{0.01\pm{0.01}}$ & $0.51\pm{0.04}$ & $80.50\pm{16.15}$ \\
 & AReS & -- & -- & -- & -- & -- & -- & -- & -- & -- \\
 & GLOBE-CE & $\boldsymbol{1.00\pm{0.00}}$ & $\boldsymbol{1.00\pm{0.00}}$ & $\boldsymbol{1.00\pm{0.00}}$ & $0.00\pm{0.00}$ & $-5.40\pm{1.77}$ & $1.86\pm{0.18}$ & $-0.07\pm{0.01}$ & $\boldsymbol{0.32\pm{0.03}}$ & $\boldsymbol{0.11\pm{0.00}}$ \\
\midrule
\multirow{3}{*}{Credit Default} & GLANCE & $\boldsymbol{1.00\pm{0.00}}$ & $0.97\pm{0.02}$ & $\boldsymbol{1.00\pm{0.00}}$ & $\boldsymbol{0.00\pm{0.00}}$ & -- & $\boldsymbol{8.49\pm{0.66}}$ & $-0.02\pm{0.02}$ & $0.86\pm{0.00}$ & $209.61\pm{15.30}$ \\
 & AReS & -- & -- & -- & -- & -- & -- & -- & -- & -- \\
 & GLOBE-CE & $\boldsymbol{1.00\pm{0.00}}$ & $\boldsymbol{1.00\pm{0.00}}$ & $0.03\pm{0.00}$ & $\boldsymbol{0.00\pm{0.00}}$ & $\boldsymbol{-957.72\pm{590.76}}$ & $10.44\pm{3.61}$ & $\boldsymbol{0.07\pm{0.00}}$ & $\boldsymbol{0.15\pm{0.04}}$ & $\boldsymbol{4.00\pm{0.10}}$ \\
\midrule
\multirow{3}{*}{Digits} & GLANCE & $\boldsymbol{1.00\pm{0.00}}$ & $\boldsymbol{0.70\pm{0.14}}$ & $\boldsymbol{0.85\pm{0.05}}$ & $\boldsymbol{0.01\pm{0.01}}$ & $\boldsymbol{-871.81\pm{845.67}}$ & $\boldsymbol{1.47\pm{0.05}}$ & $\boldsymbol{-0.07\pm{0.01}}$ & $\boldsymbol{1.55\pm{0.12}}$ & $\boldsymbol{2.14\pm{0.11}}$ \\
 & AReS & -- & -- & -- & -- & -- & -- & -- & -- & -- \\
 & GLOBE-CE & -- & $0.00\pm{0.00}$ & -- & -- & -- & -- & -- & -- & -- \\
\midrule
\multirow{3}{*}{German Credit} & GLANCE & $\boldsymbol{1.00\pm{0.00}}$ & $\boldsymbol{0.82\pm{0.14}}$ & $\boldsymbol{0.70\pm{0.19}}$ & $0.08\pm{0.11}$ & -- & $\boldsymbol{1.08\pm{0.03}}$ & $-0.04\pm{0.02}$ & $0.64\pm{0.18}$ & $2.14\pm{1.12}$ \\
 & AReS & $\boldsymbol{1.00\pm{0.00}}$ & $0.34\pm{0.14}$ & $0.04\pm{0.01}$ & $\boldsymbol{0.27\pm{0.10}}$ & $\boldsymbol{-61.66\pm{2.25}}$ & $\boldsymbol{1.08\pm{0.01}}$ & $\boldsymbol{0.01\pm{0.01}}$ & $\boldsymbol{0.08\pm{0.06}}$ & $1.99\pm{1.49}$ \\
 & GLOBE-CE & $\boldsymbol{1.00\pm{0.00}}$ & $0.66\pm{0.26}$ & $0.09\pm{0.02}$ & $0.09\pm{0.12}$ & -- & $12.13\pm{11.73}$ & $\boldsymbol{0.01\pm{0.01}}$ & $0.15\pm{0.09}$ & $\boldsymbol{0.21\pm{0.00}}$ \\
\midrule
\multirow{3}{*}{GMC} & GLANCE & $\boldsymbol{1.00\pm{0.00}}$ & $\boldsymbol{1.00\pm{0.00}}$ & $\boldsymbol{1.00\pm{0.01}}$ & $0.01\pm{0.03}$ & $-153.40\pm{291.01}$ & $5.31\pm{0.31}$ & $-0.01\pm{0.01}$ & $0.85\pm{0.01}$ & $62.40\pm{1.53}$ \\
 & AReS & $\boldsymbol{1.00\pm{0.00}}$ & $0.83\pm{0.10}$ & $0.05\pm{0.00}$ & $\boldsymbol{0.46\pm{0.23}}$ & $\boldsymbol{-12.58\pm{9.50}}$ & $1.21\pm{0.07}$ & $0.07\pm{0.03}$ & $0.05\pm{0.00}$ & $16.33\pm{6.68}$ \\
 & GLOBE-CE & $\boldsymbol{1.00\pm{0.00}}$ & $\boldsymbol{1.00\pm{0.00}}$ & $0.06\pm{0.01}$ & $0.43\pm{0.21}$ & $-14.21\pm{7.55}$ & $\boldsymbol{1.16\pm{0.05}}$ & $\boldsymbol{0.12\pm{0.01}}$ & $\boldsymbol{0.00\pm{0.00}}$ & $\boldsymbol{0.94\pm{0.01}}$ \\
\midrule
\multirow{3}{*}{HELOC} & GLANCE & $\boldsymbol{1.00\pm{0.00}}$ & $0.96\pm{0.02}$ & $\boldsymbol{1.00\pm{0.00}}$ & $0.00\pm{0.00}$ & $\boldsymbol{-543.81\pm{733.67}}$ & $6.42\times 10^{8}\pm{1.07\times 10^{8}}$ & $0.00\pm{0.01}$ & $0.61\pm{0.07}$ & $55.65\pm{5.69}$ \\
 & AReS & $\boldsymbol{1.00\pm{0.00}}$ & $0.27\pm{0.08}$ & $0.24\pm{0.01}$ & $0.07\pm{0.01}$ & $-976.16\pm{1339.37}$ & $\boldsymbol{38.07\pm{4.43}}$ & $\boldsymbol{0.04\pm{0.00}}$ & $\boldsymbol{0.20\pm{0.08}}$ & $\boldsymbol{0.15\pm{0.10}}$ \\
 & GLOBE-CE & $\boldsymbol{1.00\pm{0.00}}$ & $\boldsymbol{1.00\pm{0.00}}$ & $0.27\pm{0.01}$ & $\boldsymbol{0.10\pm{0.06}}$ & $-4402.46\pm{4600.04}$ & $1.21\times 10^{8}\pm{6.26\times 10^{7}}$ & $\boldsymbol{0.04\pm{0.00}}$ & $0.33\pm{0.12}$ & $0.33\pm{0.00}$ \\
\midrule
\multirow{3}{*}{Law} & GLANCE & $\boldsymbol{1.00\pm{0.00}}$ & $0.97\pm{0.01}$ & $\boldsymbol{1.00\pm{0.00}}$ & $\boldsymbol{0.39\pm{0.05}}$ & $\boldsymbol{-16.21\pm{1.69}}$ & $1.65\pm{0.08}$ & $0.00\pm{0.01}$ & $0.87\pm{0.01}$ & $30.29\pm{1.56}$ \\
 & AReS & -- & -- & -- & -- & -- & -- & -- & -- & -- \\
 & GLOBE-CE & $\boldsymbol{1.00\pm{0.00}}$ & $\boldsymbol{1.00\pm{0.00}}$ & $0.20\pm{0.00}$ & $0.26\pm{0.18}$ & $-16.79\pm{1.73}$ & $\boldsymbol{1.39\pm{0.06}}$ & $\boldsymbol{0.02\pm{0.01}}$ & $\boldsymbol{0.05\pm{0.00}}$ & $\boldsymbol{0.17\pm{0.00}}$ \\
\midrule
\multirow{3}{*}{Lending Club} & GLANCE & $\boldsymbol{1.00\pm{0.00}}$ & $\boldsymbol{1.00\pm{0.00}}$ & $\boldsymbol{0.99\pm{0.02}}$ & $0.00\pm{0.00}$ & $-298.92\pm{113.97}$ & $\boldsymbol{4.24\pm{0.51}}$ & $-0.03\pm{0.01}$ & $0.91\pm{0.05}$ & $64.13\pm{6.76}$ \\
 & AReS & -- & -- & -- & -- & -- & -- & -- & -- & -- \\
 & GLOBE-CE & $\boldsymbol{1.00\pm{0.00}}$ & $0.77\pm{0.22}$ & $0.18\pm{0.00}$ & $\boldsymbol{0.47\pm{0.30}}$ & $\boldsymbol{-33.06\pm{5.96}}$ & $5.39\pm{0.40}$ & $\boldsymbol{0.04\pm{0.00}}$ & $\boldsymbol{0.06\pm{0.00}}$ & $\boldsymbol{0.82\pm{0.01}}$ \\
\midrule
\multirow{3}{*}{Moons} & GLANCE & $\boldsymbol{1.00\pm{0.00}}$ & $0.99\pm{0.01}$ & $\boldsymbol{1.00\pm{0.00}}$ & $\boldsymbol{0.28\pm{0.02}}$ & $-22.19\pm{13.90}$ & $1.78\pm{0.16}$ & $-0.03\pm{0.01}$ & $0.50\pm{0.01}$ & $38.69\pm{9.27}$ \\
 & AReS & -- & -- & -- & -- & -- & -- & -- & -- & -- \\
 & GLOBE-CE & $\boldsymbol{1.00\pm{0.00}}$ & $\boldsymbol{1.00\pm{0.00}}$ & $\boldsymbol{1.00\pm{0.00}}$ & $0.00\pm{0.00}$ & $\boldsymbol{-1.26\pm{0.61}}$ & $\boldsymbol{1.31\pm{0.04}}$ & $\boldsymbol{-0.01\pm{0.01}}$ & $\boldsymbol{0.27\pm{0.02}}$ & $\boldsymbol{0.11\pm{0.00}}$ \\
\midrule
\multirow{3}{*}{Wine} & GLANCE & $\boldsymbol{1.00\pm{0.00}}$ & $0.78\pm{0.09}$ & $\boldsymbol{0.76\pm{0.18}}$ & $\boldsymbol{0.01\pm{0.04}}$ & $\boldsymbol{-26.27\pm{20.29}}$ & $\boldsymbol{1.28\pm{0.06}}$ & $\boldsymbol{0.00\pm{0.02}}$ & $\boldsymbol{0.90\pm{0.05}}$ & $1.01\pm{0.85}$ \\
 & AReS & -- & -- & -- & -- & -- & -- & -- & -- & -- \\
 & GLOBE-CE & $\boldsymbol{1.00\pm{0.00}}$ & $\boldsymbol{1.00\pm{0.00}}$ & $0.57\pm{0.04}$ & $\boldsymbol{0.01\pm{0.03}}$ & $-140.54\pm{133.50}$ & $1.79\pm{0.29}$ & $\boldsymbol{0.00\pm{0.01}}$ & $1.29\pm{0.20}$ & $\boldsymbol{0.10\pm{0.00}}$ \\
\bottomrule
\end{tabular}
\end{scriptsize}
\end{sc}
\end{center}
\end{table*}

\subsection{Group-wise Methods}

This subsection reports results for group-wise counterfactual explanation methods, which generate shared explanations for subgroups of instances. The evaluated methods include TCREx and GLANCE (GLANCE). Unlike local methods, the reported proximity and validity metrics reflect the effectiveness of shared shift vectors applied to instance subgroups rather than individually optimized perturbations. Results are reported across all 13 classification datasets and for both MLP and Logistic Regression backbones. Tables~\ref{tab:group_metrics_mlp} and~\ref{tab:group_metrics_lr} report all metrics for MLP and Logistic Regression, respectively.

TCREx exhibits limited applicability, failing to produce results on several datasets including Credit Default, Digits, German Credit, Give Me Some Credit, Lending Club, and Wine under both backbones. On datasets where it succeeds, TCREx consistently achieves the smallest perturbations, often by a substantial margin. For instance, on Adult Census it produces near-zero distances across all proximity metrics, and on HELOC it achieves L2-Hamming of 0.08 compared to 0.76 for GLANCE (MLP). However, these minimal perturbations come at the cost of extremely low validity: TCREx achieves only 0.23 on Adult Census, 0.04 on Audit, 0.01 on Blobs, and 0.03 on HELOC (MLP). This indicates that the subgroup-level shifts learned by TCREx are often too conservative to change model predictions.

In contrast, GLANCE achieves substantially higher validity across all datasets. On many configurations it reaches near-perfect validity (e.g., 1.00 on Bank Marketing, Give Me Some Credit, and Lending Club with MLP), though it shows reduced effectiveness on some datasets such as Audit (0.36 with MLP), Moons (0.48 with MLP), and Digits (0.71 with MLP). On Adult Census, GLANCE achieves 0.99 validity (MLP) compared to TCREx's 0.23, though with considerably larger perturbations. This pattern is consistent: GLANCE applies larger, more aggressive shifts that successfully flip predictions for a higher proportion of instances within each subgroup.

The trade-off between proximity and validity in the group-wise paradigm is more pronounced than in local methods. Because shared shifts must be effective across multiple instances within a subgroup, methods face a fundamental tension between minimal change and broad effectiveness. TCREx favors minimal perturbation at the expense of validity, while GLANCE prioritizes prediction flipping.

In terms of plausibility, TCREx achieves higher probabilistic plausibility and better LOF scores on datasets where it produces results (e.g., 0.49 on Adult Census, 0.26 on HELOC), reflecting the fact that its small perturbations keep counterfactuals close to the original data manifold. GLANCE generally achieves lower plausibility scores due to its larger perturbations. TCREx is also considerably faster, typically completing in under 1 second, while GLANCE typically requires tens to hundreds of seconds depending on the dataset size.

\begin{table*}[ht]
\centering
\caption{Group-wise methods: metrics (MLP).}
\label{tab:group_metrics_mlp}
\begin{center}
\begin{sc}
\begin{scriptsize}
\begin{tabular}{l|l|rrrrrrrrr}
\toprule
 & Method & Cov.$\uparrow$ & Valid.$\uparrow$ & Sparse.$\uparrow$ & Prob. Plaus.$\uparrow$ & Log Dens.$\uparrow$ & LOF$\downarrow$ & IsoForest$\uparrow$ & L2-Ham.$\downarrow$ & Time(s)$\downarrow$ \\
\midrule
\multirow{2}{*}{Adult Census} & TCREx & $\boldsymbol{1.00\pm{0.00}}$ & $0.23\pm{0.02}$ & $0.00\pm{0.00}$ & $\boldsymbol{0.49\pm{0.15}}$ & $\boldsymbol{-242.83\pm{13.92}}$ & $\boldsymbol{3.05\pm{0.05}}$ & $\boldsymbol{0.12\pm{0.00}}$ & $\boldsymbol{0.00\pm{0.00}}$ & $\boldsymbol{0.96\pm{0.03}}$ \\
 & GLANCE & $\boldsymbol{1.00\pm{0.00}}$ & $\boldsymbol{0.99\pm{0.01}}$ & $\boldsymbol{0.41\pm{0.02}}$ & $0.35\pm{0.05}$ & $-251.57\pm{80.02}$ & $3.95\pm{0.15}$ & $0.11\pm{0.00}$ & $0.41\pm{0.02}$ & $261.15\pm{1.33}$ \\
\midrule
\multirow{2}{*}{Audit} & TCREx & $\boldsymbol{1.00\pm{0.00}}$ & $0.04\pm{0.04}$ & $0.01\pm{0.01}$ & $\boldsymbol{0.10\pm{0.03}}$ & $\boldsymbol{42.81\pm{6.55}}$ & $125470.63\pm{138611.61}$ & $\boldsymbol{0.16\pm{0.01}}$ & $\boldsymbol{0.22\pm{0.09}}$ & $\boldsymbol{0.01\pm{0.00}}$ \\
 & GLANCE & $\boldsymbol{1.00\pm{0.00}}$ & $\boldsymbol{0.36\pm{0.06}}$ & $\boldsymbol{0.85\pm{0.16}}$ & $0.00\pm{0.00}$ & $7.56\pm{11.46}$ & $\boldsymbol{119.05\pm{38.01}}$ & $-0.08\pm{0.03}$ & $0.36\pm{0.03}$ & $27.69\pm{0.13}$ \\
\midrule
\multirow{2}{*}{Bank Marketing} & TCREx & $\boldsymbol{1.00\pm{0.00}}$ & $0.15\pm{0.31}$ & $0.02\pm{0.02}$ & $\boldsymbol{0.44\pm{0.29}}$ & $-47.03\pm{4.30}$ & $\boldsymbol{2.78\pm{0.64}}$ & $\boldsymbol{0.06\pm{0.01}}$ & $\boldsymbol{0.03\pm{0.02}}$ & $\boldsymbol{0.41\pm{0.01}}$ \\
 & GLANCE & $\boldsymbol{1.00\pm{0.00}}$ & $\boldsymbol{1.00\pm{0.00}}$ & $\boldsymbol{0.65\pm{0.06}}$ & $0.01\pm{0.01}$ & $\boldsymbol{0.00\pm{-\infty}}$ & $3.53\pm{0.35}$ & $0.01\pm{0.03}$ & $0.65\pm{0.05}$ & $98.64\pm{13.26}$ \\
\midrule
\multirow{2}{*}{Blobs} & TCREx & $\boldsymbol{1.00\pm{0.00}}$ & $0.01\pm{0.01}$ & $0.04\pm{0.01}$ & $0.00\pm{0.00}$ & $-40.40\pm{11.23}$ & $\boldsymbol{1.09\pm{0.02}}$ & $\boldsymbol{0.03\pm{0.01}}$ & $\boldsymbol{0.36\pm{0.12}}$ & $\boldsymbol{0.00\pm{0.00}}$ \\
 & GLANCE & $\boldsymbol{1.00\pm{0.00}}$ & $\boldsymbol{0.99\pm{0.01}}$ & $\boldsymbol{1.00\pm{0.00}}$ & $\boldsymbol{0.30\pm{0.17}}$ & $\boldsymbol{0.53\pm{0.53}}$ & $1.29\pm{0.09}$ & $-0.01\pm{0.02}$ & $0.50\pm{0.05}$ & $95.64\pm{8.00}$ \\
\midrule
\multirow{2}{*}{Credit Default} & TCREx & -- & $0.00\pm{0.00}$ & -- & -- & -- & -- & -- & -- & $\boldsymbol{0.59\pm{0.02}}$ \\
 & GLANCE & $\boldsymbol{1.00\pm{0.00}}$ & $\boldsymbol{0.98\pm{0.01}}$ & $\boldsymbol{0.92\pm{0.01}}$ & $\boldsymbol{0.00\pm{0.00}}$ & $\boldsymbol{0.00\pm{-\infty}}$ & $\boldsymbol{7.43\pm{1.71}}$ & $\boldsymbol{-0.05\pm{0.01}}$ & $\boldsymbol{0.81\pm{0.01}}$ & $235.54\pm{32.78}$ \\
\midrule
\multirow{2}{*}{Digits} & TCREx & -- & -- & -- & -- & -- & -- & -- & -- & -- \\
 & GLANCE & $\boldsymbol{1.00\pm{0.00}}$ & $\boldsymbol{0.71\pm{0.05}}$ & $\boldsymbol{0.92\pm{0.01}}$ & $\boldsymbol{0.00\pm{0.00}}$ & $\boldsymbol{-3252.85\pm{5069.03}}$ & $\boldsymbol{1.51\pm{0.02}}$ & $\boldsymbol{-0.07\pm{0.01}}$ & $\boldsymbol{1.71\pm{0.08}}$ & $\boldsymbol{2.60\pm{0.21}}$ \\
\midrule
\multirow{2}{*}{German Credit} & TCREx & -- & -- & -- & -- & -- & -- & -- & -- & -- \\
 & GLANCE & $\boldsymbol{1.00\pm{0.00}}$ & $\boldsymbol{0.83\pm{0.05}}$ & $\boldsymbol{0.90\pm{0.07}}$ & $\boldsymbol{0.30\pm{0.14}}$ & $\boldsymbol{0.00\pm{-\infty}}$ & $\boldsymbol{1.04\pm{0.01}}$ & $\boldsymbol{-0.01\pm{0.00}}$ & $\boldsymbol{0.81\pm{0.05}}$ & $\boldsymbol{88.70\pm{6.82}}$ \\
\midrule
\multirow{2}{*}{GMC} & TCREx & -- & -- & -- & -- & -- & -- & -- & -- & -- \\
 & GLANCE & $\boldsymbol{1.00\pm{0.00}}$ & $\boldsymbol{1.00\pm{0.00}}$ & $\boldsymbol{0.89\pm{0.18}}$ & $\boldsymbol{0.00\pm{0.00}}$ & $\boldsymbol{-41.83\pm{15.03}}$ & $\boldsymbol{6.08\pm{0.95}}$ & $\boldsymbol{0.01\pm{0.02}}$ & $\boldsymbol{0.76\pm{0.15}}$ & $\boldsymbol{63.09\pm{0.91}}$ \\
\midrule
\multirow{2}{*}{HELOC} & TCREx & $\boldsymbol{1.00\pm{0.00}}$ & $0.03\pm{0.00}$ & $0.03\pm{0.01}$ & $\boldsymbol{0.26\pm{0.03}}$ & $\boldsymbol{-3052.07\pm{4574.88}}$ & $\boldsymbol{1.10\pm{0.01}}$ & $\boldsymbol{0.04\pm{0.00}}$ & $\boldsymbol{0.08\pm{0.01}}$ & $\boldsymbol{0.05\pm{0.00}}$ \\
 & GLANCE & $\boldsymbol{1.00\pm{0.00}}$ & $\boldsymbol{0.96\pm{0.02}}$ & $\boldsymbol{0.78\pm{0.09}}$ & $0.00\pm{0.00}$ & $-99808.75\pm{219402.22}$ & $1.05\times 10^{9}\pm{1.65\times 10^{8}}$ & $0.00\pm{0.01}$ & $0.76\pm{0.04}$ & $66.65\pm{7.95}$ \\
\midrule
\multirow{2}{*}{Law} & TCREx & $\boldsymbol{1.00\pm{0.00}}$ & $0.04\pm{0.05}$ & $0.04\pm{0.02}$ & $0.02\pm{0.02}$ & $-20.59\pm{3.21}$ & $\boldsymbol{1.29\pm{0.01}}$ & $\boldsymbol{0.03\pm{0.01}}$ & $\boldsymbol{0.07\pm{0.03}}$ & $\boldsymbol{0.01\pm{0.00}}$ \\
 & GLANCE & $\boldsymbol{1.00\pm{0.00}}$ & $\boldsymbol{0.94\pm{0.02}}$ & $\boldsymbol{0.92\pm{0.13}}$ & $\boldsymbol{0.41\pm{0.08}}$ & $\boldsymbol{-12.68\pm{1.53}}$ & $1.81\pm{0.37}$ & $0.00\pm{0.01}$ & $0.80\pm{0.13}$ & $27.19\pm{3.96}$ \\
\midrule
\multirow{2}{*}{Lending Club} & TCREx & -- & -- & -- & -- & -- & -- & -- & -- & -- \\
 & GLANCE & $\boldsymbol{1.00\pm{0.00}}$ & $\boldsymbol{1.00\pm{0.00}}$ & $\boldsymbol{0.84\pm{0.12}}$ & $\boldsymbol{0.03\pm{0.06}}$ & $\boldsymbol{0.00\pm{-\infty}}$ & $\boldsymbol{3.22\pm{0.24}}$ & $\boldsymbol{-0.04\pm{0.02}}$ & $\boldsymbol{0.71\pm{0.11}}$ & $\boldsymbol{82.73\pm{12.03}}$ \\
\midrule
\multirow{2}{*}{Moons} & TCREx & $\boldsymbol{1.00\pm{0.00}}$ & $0.10\pm{0.01}$ & $0.25\pm{0.03}$ & $0.00\pm{0.00}$ & $-39.46\pm{26.23}$ & $\boldsymbol{1.22\pm{0.02}}$ & $0.01\pm{0.01}$ & $\boldsymbol{0.17\pm{0.01}}$ & $\boldsymbol{0.00\pm{0.00}}$ \\
 & GLANCE & $\boldsymbol{1.00\pm{0.00}}$ & $\boldsymbol{0.48\pm{0.09}}$ & $\boldsymbol{1.00\pm{0.00}}$ & $\boldsymbol{0.12\pm{0.04}}$ & $\boldsymbol{-6.22\pm{2.38}}$ & $1.25\pm{0.03}$ & $\boldsymbol{0.02\pm{0.01}}$ & $0.44\pm{0.05}$ & $43.71\pm{9.73}$ \\
\midrule
\multirow{2}{*}{Wine} & TCREx & -- & -- & -- & -- & -- & -- & -- & -- & -- \\
 & GLANCE & $\boldsymbol{1.00\pm{0.00}}$ & $\boldsymbol{0.79\pm{0.05}}$ & $\boldsymbol{0.78\pm{0.10}}$ & $\boldsymbol{0.04\pm{0.06}}$ & $\boldsymbol{-18.74\pm{7.78}}$ & $\boldsymbol{1.23\pm{0.03}}$ & $\boldsymbol{0.00\pm{0.01}}$ & $\boldsymbol{0.90\pm{0.06}}$ & $\boldsymbol{0.50\pm{0.04}}$ \\
\bottomrule
\end{tabular}
\end{scriptsize}
\end{sc}
\end{center}
\end{table*}

\begin{table*}[ht]
\centering
\caption{Group-wise methods: metrics (LR).}
\label{tab:group_metrics_lr}
\begin{center}
\begin{sc}
\begin{scriptsize}
\begin{tabular}{l|l|rrrrrrrrr}
\toprule
 & Method & Cov.$\uparrow$ & Valid.$\uparrow$ & Sparse.$\uparrow$ & Prob. Plaus.$\uparrow$ & Log Dens.$\uparrow$ & LOF$\downarrow$ & IsoForest$\uparrow$ & L2-Ham.$\downarrow$ & Time(s)$\downarrow$ \\
\midrule
\multirow{2}{*}{Adult Census} & TCREx & $\boldsymbol{1.00\pm{0.00}}$ & $0.23\pm{0.02}$ & $0.00\pm{0.00}$ & $\boldsymbol{0.36\pm{0.18}}$ & $-249.50\pm{11.74}$ & $\boldsymbol{3.04\pm{0.05}}$ & $\boldsymbol{0.12\pm{0.00}}$ & $\boldsymbol{0.00\pm{0.00}}$ & $\boldsymbol{0.96\pm{0.01}}$ \\
 & GLANCE & $\boldsymbol{1.00\pm{0.00}}$ & $\boldsymbol{1.00\pm{0.00}}$ & $\boldsymbol{0.40\pm{0.03}}$ & $0.35\pm{0.05}$ & $\boldsymbol{-141.41\pm{57.06}}$ & $4.08\pm{0.19}$ & $0.11\pm{0.00}$ & $0.40\pm{0.03}$ & $258.84\pm{1.50}$ \\
\midrule
\multirow{2}{*}{Audit} & TCREx & $\boldsymbol{1.00\pm{0.00}}$ & $0.01\pm{0.00}$ & $0.06\pm{0.09}$ & $\boldsymbol{0.13\pm{0.17}}$ & $\boldsymbol{30.61\pm{23.07}}$ & $16847.39\pm{22812.53}$ & $\boldsymbol{0.15\pm{0.02}}$ & $\boldsymbol{0.15\pm{0.18}}$ & $\boldsymbol{0.01\pm{0.00}}$ \\
 & GLANCE & $\boldsymbol{1.00\pm{0.00}}$ & $\boldsymbol{1.00\pm{0.00}}$ & $\boldsymbol{0.94\pm{0.11}}$ & $0.00\pm{0.00}$ & $15.70\pm{8.24}$ & $\boldsymbol{98.44\pm{27.47}}$ & $-0.09\pm{0.04}$ & $0.28\pm{0.01}$ & $27.59\pm{0.11}$ \\
\midrule
\multirow{2}{*}{Bank Marketing} & TCREx & $\boldsymbol{1.00\pm{0.00}}$ & $0.01\pm{0.01}$ & $0.01\pm{0.00}$ & $\boldsymbol{0.20\pm{0.18}}$ & $-51.93\pm{5.64}$ & $\boldsymbol{2.31\pm{0.55}}$ & $\boldsymbol{0.06\pm{0.00}}$ & $\boldsymbol{0.03\pm{0.01}}$ & $\boldsymbol{0.40\pm{0.00}}$ \\
 & GLANCE & $\boldsymbol{1.00\pm{0.00}}$ & $\boldsymbol{1.00\pm{0.00}}$ & $\boldsymbol{0.58\pm{0.03}}$ & $0.16\pm{0.10}$ & $\boldsymbol{0.00\pm{-\infty}}$ & $2.77\pm{0.05}$ & $0.03\pm{0.00}$ & $0.57\pm{0.03}$ & $91.52\pm{2.32}$ \\
\midrule
\multirow{2}{*}{Blobs} & TCREx & $\boldsymbol{1.00\pm{0.00}}$ & $0.01\pm{0.01}$ & $0.03\pm{0.01}$ & $0.00\pm{0.00}$ & $-39.68\pm{18.99}$ & $\boldsymbol{1.08\pm{0.03}}$ & $\boldsymbol{0.03\pm{0.01}}$ & $\boldsymbol{0.37\pm{0.17}}$ & $\boldsymbol{0.00\pm{0.00}}$ \\
 & GLANCE & $\boldsymbol{1.00\pm{0.00}}$ & $\boldsymbol{1.00\pm{0.00}}$ & $\boldsymbol{1.00\pm{0.00}}$ & $\boldsymbol{0.43\pm{0.13}}$ & $\boldsymbol{0.89\pm{0.39}}$ & $1.22\pm{0.05}$ & $0.00\pm{0.01}$ & $0.53\pm{0.05}$ & $80.43\pm{14.75}$ \\
\midrule
\multirow{2}{*}{Credit Default} & TCREx & -- & $0.00\pm{0.00}$ & -- & -- & -- & -- & -- & -- & $\boldsymbol{0.56\pm{0.02}}$ \\
 & GLANCE & $\boldsymbol{1.00\pm{0.00}}$ & $\boldsymbol{0.97\pm{0.02}}$ & $\boldsymbol{0.86\pm{0.03}}$ & $\boldsymbol{0.00\pm{0.00}}$ & $\boldsymbol{0.00\pm{-\infty}}$ & $\boldsymbol{7.51\pm{1.78}}$ & $\boldsymbol{-0.03\pm{0.02}}$ & $\boldsymbol{0.76\pm{0.02}}$ & $195.49\pm{12.75}$ \\
\midrule
\multirow{2}{*}{Digits} & TCREx & -- & -- & -- & -- & -- & -- & -- & -- & -- \\
 & GLANCE & $\boldsymbol{1.00\pm{0.00}}$ & $\boldsymbol{0.73\pm{0.09}}$ & $\boldsymbol{0.81\pm{0.03}}$ & $\boldsymbol{0.00\pm{0.00}}$ & $\boldsymbol{-1050.00\pm{849.44}}$ & $\boldsymbol{1.47\pm{0.03}}$ & $\boldsymbol{-0.07\pm{0.01}}$ & $\boldsymbol{1.57\pm{0.09}}$ & $\boldsymbol{2.14\pm{0.13}}$ \\
\midrule
\multirow{2}{*}{German Credit} & TCREx & -- & -- & -- & -- & -- & -- & -- & -- & -- \\
 & GLANCE & $\boldsymbol{1.00\pm{0.00}}$ & $\boldsymbol{0.82\pm{0.12}}$ & $\boldsymbol{0.92\pm{0.05}}$ & $\boldsymbol{0.24\pm{0.16}}$ & $\boldsymbol{0.00\pm{-\infty}}$ & $\boldsymbol{1.05\pm{0.01}}$ & $\boldsymbol{-0.02\pm{0.02}}$ & $\boldsymbol{0.84\pm{0.05}}$ & $\boldsymbol{81.75\pm{13.45}}$ \\
\midrule
\multirow{2}{*}{GMC} & TCREx & -- & -- & -- & -- & -- & -- & -- & -- & -- \\
 & GLANCE & $\boldsymbol{1.00\pm{0.00}}$ & $\boldsymbol{1.00\pm{0.00}}$ & $\boldsymbol{0.97\pm{0.01}}$ & $\boldsymbol{0.00\pm{0.00}}$ & $\boldsymbol{-143.36\pm{261.87}}$ & $\boldsymbol{5.34\pm{0.31}}$ & $\boldsymbol{-0.01\pm{0.02}}$ & $\boldsymbol{0.83\pm{0.01}}$ & $\boldsymbol{62.09\pm{0.94}}$ \\
\midrule
\multirow{2}{*}{HELOC} & TCREx & $\boldsymbol{1.00\pm{0.00}}$ & $0.03\pm{0.01}$ & $0.03\pm{0.00}$ & $\boldsymbol{0.26\pm{0.03}}$ & $-1585.94\pm{1307.54}$ & $\boldsymbol{1.10\pm{0.00}}$ & $\boldsymbol{0.04\pm{0.00}}$ & $\boldsymbol{0.09\pm{0.01}}$ & $\boldsymbol{0.04\pm{0.00}}$ \\
 & GLANCE & $\boldsymbol{1.00\pm{0.00}}$ & $\boldsymbol{0.95\pm{0.04}}$ & $\boldsymbol{0.72\pm{0.08}}$ & $0.00\pm{0.00}$ & $\boldsymbol{-194.78\pm{134.95}}$ & $5.65\times 10^{8}\pm{1.35\times 10^{8}}$ & $0.00\pm{0.00}$ & $0.62\pm{0.11}$ & $54.01\pm{3.18}$ \\
\midrule
\multirow{2}{*}{Law} & TCREx & $\boldsymbol{1.00\pm{0.00}}$ & $0.06\pm{0.05}$ & $0.05\pm{0.01}$ & $0.15\pm{0.14}$ & $-16.73\pm{1.53}$ & $\boldsymbol{1.31\pm{0.08}}$ & $\boldsymbol{0.03\pm{0.01}}$ & $\boldsymbol{0.06\pm{0.03}}$ & $\boldsymbol{0.01\pm{0.00}}$ \\
 & GLANCE & $\boldsymbol{1.00\pm{0.00}}$ & $\boldsymbol{0.97\pm{0.01}}$ & $\boldsymbol{0.99\pm{0.00}}$ & $\boldsymbol{0.40\pm{0.06}}$ & $\boldsymbol{-15.83\pm{1.50}}$ & $1.69\pm{0.08}$ & $0.00\pm{0.00}$ & $0.87\pm{0.01}$ & $27.55\pm{2.14}$ \\
\midrule
\multirow{2}{*}{Lending Club} & TCREx & -- & -- & -- & -- & -- & -- & -- & -- & -- \\
 & GLANCE & $\boldsymbol{1.00\pm{0.00}}$ & $\boldsymbol{1.00\pm{0.00}}$ & $\boldsymbol{0.68\pm{0.11}}$ & $\boldsymbol{0.00\pm{0.00}}$ & $\boldsymbol{-44.69\pm{-\infty}}$ & $\boldsymbol{4.13\pm{0.27}}$ & $\boldsymbol{-0.03\pm{0.00}}$ & $\boldsymbol{0.67\pm{0.09}}$ & $\boldsymbol{65.58\pm{5.62}}$ \\
\midrule
\multirow{2}{*}{Moons} & TCREx & $\boldsymbol{1.00\pm{0.00}}$ & $0.13\pm{0.06}$ & $0.38\pm{0.06}$ & $0.06\pm{0.02}$ & $-1236.33\pm{1652.67}$ & $\boldsymbol{1.17\pm{0.10}}$ & $-0.01\pm{0.03}$ & $\boldsymbol{0.49\pm{0.00}}$ & $\boldsymbol{0.00\pm{0.00}}$ \\
 & GLANCE & $\boldsymbol{1.00\pm{0.00}}$ & $\boldsymbol{0.99\pm{0.01}}$ & $\boldsymbol{1.00\pm{0.00}}$ & $\boldsymbol{0.41\pm{0.15}}$ & $\boldsymbol{-0.04\pm{0.64}}$ & $1.29\pm{0.05}$ & $\boldsymbol{0.00\pm{0.01}}$ & $0.52\pm{0.01}$ & $34.16\pm{3.77}$ \\
\midrule
\multirow{2}{*}{Wine} & TCREx & -- & -- & -- & -- & -- & -- & -- & -- & -- \\
 & GLANCE & $\boldsymbol{1.00\pm{0.00}}$ & $\boldsymbol{0.78\pm{0.07}}$ & $\boldsymbol{0.73\pm{0.14}}$ & $\boldsymbol{0.00\pm{0.00}}$ & $\boldsymbol{-21.38\pm{11.80}}$ & $\boldsymbol{1.27\pm{0.07}}$ & $\boldsymbol{0.00\pm{0.02}}$ & $\boldsymbol{0.92\pm{0.05}}$ & $\boldsymbol{0.74\pm{0.69}}$ \\
\bottomrule
\end{tabular}
\end{scriptsize}
\end{sc}
\end{center}
\end{table*}

\bibliographystyle{splncs04}
\bibliography{sample-base}